\PassOptionsToPackage{unicode}{hyperref}
\PassOptionsToPackage{hyphens}{url}
\PassOptionsToPackage{dvipsnames,svgnames,x11names}{xcolor}
\documentclass[
  letterpaper,
  DIV=11,
  numbers=noendperiod]{scrartcl}
\usepackage{xcolor}
\usepackage{amsmath,amssymb}
\usepackage{iftex}
\ifPDFTeX
  \usepackage[T1]{fontenc}
  \usepackage[utf8]{inputenc}
  \usepackage{textcomp} 
\else 
  \usepackage{unicode-math} 
  \defaultfontfeatures{Scale=MatchLowercase}
  \defaultfontfeatures[\rmfamily]{Ligatures=TeX,Scale=1}
\fi
\usepackage{lmodern}
\ifPDFTeX\else
\fi
\IfFileExists{upquote.sty}{\usepackage{upquote}}{}
\IfFileExists{microtype.sty}{
  \usepackage[]{microtype}
  \UseMicrotypeSet[protrusion]{basicmath} 
}{}
\makeatletter
\@ifundefined{KOMAClassName}{
  \IfFileExists{parskip.sty}{%
    \usepackage{parskip}
  }{
    \setlength{\parindent}{0pt}
    \setlength{\parskip}{6pt plus 2pt minus 1pt}}
}{
  \KOMAoptions{parskip=half}}
\makeatother
\makeatletter
\ifx\paragraph\undefined\else
  \let\oldparagraph\paragraph
  \renewcommand{\paragraph}{
    \@ifstar
      \xxxParagraphStar
      \xxxParagraphNoStar
  }
  \newcommand{\xxxParagraphStar}[1]{\oldparagraph*{#1}\mbox{}}
  \newcommand{\xxxParagraphNoStar}[1]{\oldparagraph{#1}\mbox{}}
\fi
\ifx\subparagraph\undefined\else
  \let\oldsubparagraph\subparagraph
  \renewcommand{\subparagraph}{
    \@ifstar
      \xxxSubParagraphStar
      \xxxSubParagraphNoStar
  }
  \newcommand{\xxxSubParagraphStar}[1]{\oldsubparagraph*{#1}\mbox{}}
  \newcommand{\xxxSubParagraphNoStar}[1]{\oldsubparagraph{#1}\mbox{}}
\fi
\makeatother

\usepackage{longtable,booktabs,array}
\usepackage{calc} 
\usepackage{etoolbox}
\makeatletter
\patchcmd\longtable{\par}{\if@noskipsec\mbox{}\fi\par}{}{}
\makeatother
\IfFileExists{footnotehyper.sty}{\usepackage{footnotehyper}}{\usepackage{footnote}}
\makesavenoteenv{longtable}
\usepackage{graphicx}
\makeatletter
\newsavebox\pandoc@box
\newcommand*\pandocbounded[1]{
  \sbox\pandoc@box{#1}%
  \Gscale@div\@tempa{\textheight}{\dimexpr\ht\pandoc@box+\dp\pandoc@box\relax}%
  \Gscale@div\@tempb{\linewidth}{\wd\pandoc@box}%
  \ifdim\@tempb\p@<\@tempa\p@\let\@tempa\@tempb\fi
  \ifdim\@tempa\p@<\p@\scalebox{\@tempa}{\usebox\pandoc@box}%
  \else\usebox{\pandoc@box}%
  \fi%
}
\def\fps@figure{htbp}
\makeatother

\NewDocumentCommand\citeproctext{}{}

\makeatletter
 \let\@cite@ofmt\@firstofone
 \def\@biblabel#1{}
 \def\@cite#1#2{{#1\if@tempswa , #2\fi}}
\makeatother
\newlength{\cslhangindent}
\newlength{\csllabelwidth}
\newenvironment{CSLReferences}[2] 
 {\begin{list}{}{%
  \setlength{\itemindent}{0pt}
  \setlength{\leftmargin}{0pt}
  \setlength{\parsep}{0pt}
  \ifodd #1
   \setlength{\leftmargin}{\cslhangindent}
   \setlength{\itemindent}{-1\cslhangindent}
  \fi
  \setlength{\itemsep}{#2\baselineskip}}}
 {\end{list}}
\usepackage{calc}

\usepackage{float}
\usepackage{tabularray}
\usepackage[normalem]{ulem}
\usepackage{graphicx}
\usepackage{rotating}
\UseTblrLibrary{siunitx}
\NewTableCommand{\tinytableDefineColor}[3]{\definecolor{#1}{#2}{#3}}

\usepackage{float}
\usepackage{epigraph}
\KOMAoption{captions}{tableheading}
\makeatletter
\@ifpackageloaded{caption}{}{\usepackage{caption}}
\AtBeginDocument{%
\ifdefined\contentsname
  \renewcommand*\contentsname{Table of contents}
\else
  \newcommand\contentsname{Table of contents}
\fi
\ifdefined\listfigurename
  \renewcommand*\listfigurename{List of Figures}
\else
  \newcommand\listfigurename{List of Figures}
\fi
\ifdefined\listtablename
  \renewcommand*\listtablename{List of Tables}
\else
  \newcommand\listtablename{List of Tables}
\fi
\ifdefined\figurename
  \renewcommand*\figurename{Figure}
\else
  \newcommand\figurename{Figure}
\fi
\ifdefined\tablename
  \renewcommand*\tablename{Table}
\else
  \newcommand\tablename{Table}
\fi
}
\@ifpackageloaded{float}{}{\usepackage{float}}
\floatstyle{ruled}
\@ifundefined{c@chapter}{\newfloat{codelisting}{h}{lop}}{\newfloat{codelisting}{h}{lop}[chapter]}
\floatname{codelisting}{Listing}

\makeatother
\makeatletter
\@ifpackageloaded{caption}{}{\usepackage{caption}}
\@ifpackageloaded{subcaption}{}{\usepackage{subcaption}}
\makeatother
\usepackage{bookmark}
\IfFileExists{xurl.sty}{\usepackage{xurl}}{} 
\hypersetup{
  pdftitle={GardenBench: A lightweight{,} daily evaluation of LLM capabilities},
  pdfauthor={Benedict Cummins-Mburu\^{}\textbackslash dagger; Mariana Garcia Mejia\^{}\textbackslash dagger; Oscar Heath\^{}\textbackslash dagger; Sabrina Kreyzerman\^{}\textbackslash dagger; Ellie Murray\^{}\textbackslash dagger; Arusan Surendiran; Rohan Alexander},
  colorlinks=true,
  linkcolor={blue},
  filecolor={Maroon},
  citecolor={Blue},
  urlcolor={Blue},
  pdfcreator={LaTeX via pandoc}}

\title{\texttt{GardenBench}: A lightweight, daily evaluation of LLM
capabilities\thanks{The live dashboard is available at:
\url{https://gardenbench.ai/}. We gratefully acknowledge funding from
Natural Sciences and Engineering Research Council of Canada (NSERC)
Alliance Grant (110\_2024\_2025\_Q4\_13) and Social Sciences and
Humanities Research Council (SSHRC) Partnership Grant (895-2025-1002).
This research was conducted while Benedict Cummins-Mburu, Mariana Garcia
Mejia, Oscar Heath, Sabrina Kreyzerman, Ellie Murray and Arusan
Surendiran were interns at the Investigative Journalism Foundation. We
thank Zane Schwartz for comments and suggestions. Correspondence:
\href{mailto:rohan.alexander@utoronto.ca}{\nolinkurl{rohan.alexander@utoronto.ca}}.
\newline \(\dagger\) Equal contribution. \newline}}
\author{Benedict Cummins-Mburu\(^\dagger\) \and Mariana Garcia
Mejia\(^\dagger\) \and Oscar Heath\(^\dagger\) \and Sabrina
Kreyzerman\(^\dagger\) \and Ellie Murray\(^\dagger\) \and Arusan
Surendiran \and Rohan Alexander}
\date{September 7, 2026}
\begin{document}
\maketitle
\begin{abstract}
We introduce \texttt{GardenBench}, a benchmark with 11 test suites that
evaluate the capabilities of Large Language Models (LLMs) across a range
of tasks. By systematically prompting LLMs with various lightweight
queries, we document day-to-day changes in model performance, including
run costs and times, that other benchmarks miss. This allows us to
compare the accuracy and features of responses across model providers
and versions. We account for daily variations in model accuracy,
capturing both model progress and degradation, for instance, when a
model is changed post-deployment or when a new model is released. We
base our evaluation on general-use tasks and make our results publicly
accessible via a live dashboard, diversifying existing approaches to LLM
testing. Our main contribution is an innovative, lightweight, and
inexpensive benchmark, designed to be consumer--facing and for everyday
tasks.\\

\end{abstract}

\epigraph{\textit{One must cultivate one's own garden.}}{---Voltaire}

\section{Introduction}\label{introduction}

Progress depends on the instruments we use to measure it. Traditionally,
machine learning (ML) models were trained for and tested on
task-specific functions. For example, models were deliberately trained
to perform visual recognition and object classification in the ImageNet
challenge (Russakovsky et al. 2015) and movie recommendation in the
Netflix Prize contest (Bell et al. 2010). However, in 2020, the release
of GPT-3 (Brown et al. 2020), one of the first publicly accessible and
general-use large transformer models, shifted the trajectory of LLM
training and testing. Brown et al. (2020) found that sufficiently large
transformer models that were trained only on next-token prediction could
perform competitively with fine-tuned state-of-the-art (SOTA) models on
tasks that were not in their training data. This property of LLMs drove
their rapid spread as general-purpose tools. Today, LLMs are used as
search engines, office assistants, and personal tutors. LLMs are
referred to as foundation models (Bommasani et al. 2022; Liang et al.
2023), used to perform a broad range of tasks, including text
summarization, translation, coding, image generation, and answering
practical questions.

Though LLM use has broadened, the evaluation and documentation of LLM
performance remains scientifically oriented and domain-specific. In
addition, many model benchmarks are computationally expensive and take
long to run (Polo et al. 2024). As a result, model providers assess
overall performance by evaluating newly released models once upon their
release. Model capabilities across specialized tasks are reported in
model cards, such as GPT-5.5 (OpenAI 2026a), Gemini 3.5 Flash (Google
DeepMind 2026), and Claude Opus 5 (Anthropic 2026b).

We identify two shortcomings in this practice. First, model cards are
published by model developers, allowing them to selectively showcase
model abilities and exclude failures. The same applies for many
benchmarks, with model developers either choosing which benchmark
performances to publish or constructing their own (for example,
Anthropic (2026a)'s \texttt{BioMysteryBench} to evaluate Claude Fable
5). Second, early one-time evaluation of models does not capture changes
in performance and reliability over time. Model drift refers to the
performance degradation of LLMs as data and language patterns routinely
change (Nicholson 2026). Without running benchmarks repeatedly over
time, we are unable to quantify or record these changes properly.
Therefore, we emphasize the need for cost-effective, lightweight
benchmarks that can run every day.

In this paper, we respond to calls for the adoption of publicly
accessible benchmarks (Guha et al. 2026; Blagec et al. 2023), as well as
benchmarks that evaluate model performance more frequently and on
diverse and general use cases (for example, Hodak et al. 2024; Liang et
al. 2023). This kind of wide-ranging evaluation is intended to align LLM
benchmarks with broader LLM use.

We introduce \texttt{GardenBench}, a lightweight, fast, and inexpensive
LLM benchmark designed to run daily on all frontier models.
\texttt{GardenBench} consists of 105 tests organized into 11 suites with
varying tasks and difficulty levels. Like a garden, the benchmark
represents diversity and variety---in terms of the nature of its tasks
and its tested capabilities---within one environment. We use
Internet-sourced images, author-collected photos, and standardized text
prompts to evaluate model performance against exact-match strings on
deterministic ground truths. We test frontier models from Anthropic,
DeepSeek, OpenAI, Google, Moonshot AI, Alibaba Cloud, and Cohere by
running our benchmark once per day. This process was automated using the
GitHub Actions workflow outlined in De Angelis (2026).

We expand on existing benchmarks in a number of ways. Primarily, our
tests include a combination of innovative and practical tasks. Whereas
many benchmarks focus on specific skills, such as coding (SWE-bench
(Jimenez et al. 2024); Terminal-Bench (Merrill et al. 2026)),
summarization (SummEval (Fabbri et al. 2021); UserSumBench (Wang et al.
2024)) and translation (LingualX64 (Song et al. 2025); MMLU-ProX (Xuan
et al. 2025)), our tests measure model capability across tasks that
require multiple skills, such as image recognition, logical and
arithmetic reasoning, and information extraction. Second, the complete
benchmark costs less than \$50 USD in total to run, allowing for
personal use by interested parties. Third, our tests run every day and
our results are published daily to a publicly accessible live
dashboard\footnote{\url{https://gardenbench.ai/}}.

For the purpose of this paper, we explore results obtained from running
\texttt{GardenBench} daily from August 17, 2026 to August 31, 2026. We
begin by directly comparing model performance, first by grouping
accuracy by model, then by test suite. We then explore the relationship
between model performance and practical usage considerations, including
run cost, time spent on tasks, and token usage. Lastly, we analyze model
accuracy over time, finding that model performance varies meaningfully
day to day. Fluctuations over time were substantial enough to affect the
rankings of the models against our benchmark. Furthermore, some models
performed more variably than others. For instance, the three most
performant models (GPT-5.6 Sol, GPT-5.5, Claude Fable 5) frequently swap
between podium placements.

The remainder of this paper is structured as follows.
Section~\ref{sec-background} provides a background where we address the
objectives of \texttt{GardenBench} informed by existing benchmarks and
literature. Section~\ref{sec-method} details our methodological
approach, including the construction of our dataset and the models that
are evaluated. Section~\ref{sec-results} covers our findings regarding
model performance against metrics such as run cost and time.
Section~\ref{sec-discussion} discusses key takeaways, implications,
weaknesses, and future directions.

\section{Background}\label{sec-background}

In machine learning (ML), standard benchmarks are measurement
instruments consisting of queries with corresponding ground truth
answers and an evaluation metric for model performance across different
tasks (Raji et al. 2021). Traditionally, LLM benchmarks are designed to
test specific capabilities, such as summarizing documents
(\texttt{SummEval} (Fabbri et al. 2021); \texttt{UserSumBench} (Wang et
al. 2024)) and coding (\texttt{SWE-bench} (Jimenez et al. 2024);
\texttt{Terminal-Bench} (Merrill et al. 2026)). Other LLM benchmarks are
tailored to specific domains of knowledge, such as math
(\texttt{MathBench} (Liu et al. 2024)) and biology (\texttt{LABBench2}
(Laurent et al. 2026)), or to general use cases (\texttt{MMLU}
(Hendrycks et al. 2021); \texttt{GPQA} (Rein et al. 2023)). LLM
benchmarks also intend to compare models through objective metrics (Raji
et al. 2021), allowing for the quantitative tracking of LLM progress,
the ranking of models by performance, and the identification of the most
promising technologies to guide research efforts and general usage (Zhu
et al. 2025).

\texttt{GardenBench} follows existing publicly accessible (Chiang et al.
2024; Epoch AI 2026) and general use (Hendrycks et al. 2021; Rein et al.
2023) benchmarks that address two key concerns in LLM benchmarking.
First, LLM benchmarks tend to be highly specialized, making results less
useful to the broader public and dominated by field professionals (Koch
et al. 2021). Second, evaluation by general-use benchmarks is often
costly, meaning these tests are typically only run once upon the release
of a new model.

\subsection{Transparency and reliability concerns in domain-specific
benchmarks}\label{transparency-and-reliability-concerns-in-domain-specific-benchmarks}

\subsubsection{Selective reporting}\label{selective-reporting}

Concerns about transparency in LLM benchmarks are well documented and
characterize several important considerations. LLM evaluation occurs
largely as a closed-door practice, where model providers often
selectively publish their own evaluations (Cheng et al. 2025; Baack et
al. 2026), benchmark results target niche professional communities (Koch
et al. 2021), and evaluation metrics are often not useful and difficult
to interpret for the broad scope of LLM users (Raji et al. 2021;
Eriksson et al. 2025). For example, Guha et al. (2026) calls for more
publicly accessible and transparent LLM benchmarks in law. They argue
that most LLM benchmarks either publish all the code and data used,
which risks data contamination and may make results misleading or
difficult to interpret, or selectively report findings and discuss them
in an overly broad manner. Both approaches undermine transparency.

Conventionally, newly released ML models are evaluated on a set of
highly specialized benchmarks, with the results used to demonstrate the
model's strengths. This enables providers to publish benchmarks on which
their models perform well, while less successful performances remain
undisclosed (Cheng et al. 2025; Baack et al. 2026). For instance, party
researchers have reported lower scores for models on certain benchmarks
compared to scores reported by model creators. Akter et al. (2023)
reported a 5-shot MMLU score of 65.22 percent, whereas the Gemini paper
reported a 5-shot MMLU score of 71.8 percent (Gemini Team 2023). Efforts
to address this problem include Bommasani et al. (2023), who introduced
the Foundation Model Transparency Index (FMTI) in 2023 to expose and
improve the disclosure practices of frontier model developers (for the
most recent version, see Wan et al. (2025)). The FMTI intends to advance
the transparency of model developers across their computational methods,
labor practices, the real-world implications of LLMs, and other metrics.

\subsubsection{Specialization and professional
communities}\label{specialization-and-professional-communities}

AI companies test their new models in a set of widely used and
domain-specific benchmarks. For example, when Anthropic released Claude
Fable 5 in July 2026, the company highlighted results from
\texttt{SWE-bench} (Jimenez et al. 2024), \texttt{Terminal-Bench}
(Merrill et al. 2026), \texttt{ExploitBench} (Lee and Brumley 2026), and
\texttt{BioMysteryBench} (Anthropic 2026a). \texttt{BioMysteryBench} in
particular was developed by Anthropic to test its own model.
Respectively, these benchmarks test agentic coding, command-line coding,
cybersecurity, and biology. GPT-5.6 Sol was also evaluated on similar
measures (see Reuel et al. 2024; Li and Ho 2026; Lee and Brumley 2026;
Wang et al. 2026). While these insights are important for model
development and medical applications, these metrics may not be useful to
LLM users beyond these professional communities, who use LLMs for
ordinary daily tasks.

It has become increasingly difficult for the general public to interpret
benchmark results as they are often highly technical and
domain-specific. Guha et al. (2026) argues that the capacity to
understand and apply benchmark information largely depends on the
technical sophistication of individuals in a specific community, making
professionals who are more experienced in using AI more likely to
benefit from benchmark reports.

There is also an important discrepancy between LLM potential and real
LLM applications, likely due to a misalignment between capabilities
demonstrated by model developers and real-world applications (Narayanan
2026; Eriksson et al. 2025; Hutchinson et al. 2022; Cheng et al. 2025).
Moreover, this may also be because current approaches to LLM evaluation
may not measure what is relevant or useful to the broader community of
LLM users. Many benchmarks may fall short in capturing real-world use
cases (Eriksson et al. 2025). For instance, OpenAI et al. (2024)'s
Technical Report evaluated the model's performance on the bar exam, but
this does not reflect the everyday tasks of legal professionals
(Narayanan and Kapoor 2023). Narayanan and Kapoor (2023)'s example
highlights the gap between LLM benchmark tests and real-world utility.
Eriksson et al. (2025)'s comprehensive literature review of benchmarking
practices addresses other important vulnerabilities of current
approaches, including that many benchmarks are ill-suited for practical
use in the real world and display a serious lack of documentation. They
have also been criticized for being too narrow and primarily text-based,
ignoring capabilities of models in other modalities such as imagery and
sound.

These features of existing LLM testing support the need for
general-purpose benchmarks that are publicly accessible, trustworthy,
and interpretable. They must also indiscriminately test and publish a
wide range of capabilities across text- and image-based tasks that are
important to broad LLM users.

\subsubsection{Data contamination}\label{data-contamination}

Lastly, data contamination, which happens when test data leaks into
models' training data and leads models to perform exceptionally well
(Dong et al. 2024), may also undermine the trustworthiness of LLM
benchmarks. Cheng et al. (2025) highlights that data contamination can
lead to biased model evaluations. This makes it difficult to distinguish
between exaggerated claims about model performance and genuine progress.

\subsection{The limitations of expensive model
testing}\label{the-limitations-of-expensive-model-testing}

\subsubsection{Expensive model testing}\label{expensive-model-testing}

The expensive cost of many LLM benchmarks is another vulnerability.
Gundlach et al. (2026) found that the cost of benchmarking frontier
models is increasing by 3 to 18 times yearly, as models continue to grow
in size and require more internal thinking before outputting responses.
As a result, running the same benchmarks is becoming more costly each
time as newly released models and benchmarks themselves become more
expensive to run. One example is Kapoor et al. (2025)'s ``Holistic
Agentic Leaderboard'' which cost \$40,000 USD to run once across nine
models and nine benchmarks. GPT-6 Astra tests on ARC-AGI-3 cost \$26,000
USD at its most efficient reasoning (Kamradt 2026). Lastly, the General
AI Assistants (GAIA) benchmark (Mialon et al. 2023) cost over \$2,829
USD on frontier models for a single run (Ndzomga 2026).

\subsubsection{Static benchmarking}\label{static-benchmarking}

The increasing cost of running benchmarks and the need for frequent
evaluation are at odds with one another. The cost of evaluating LLMs
creates a new barrier that limits how frequently LLMs can be tested.
Scholars have brought attention to the computational resources needed to
evaluate LLM performance, taking realistic economic constraints into
account when assessing the feasibility of running benchmarks to capture
reliability and other characteristics of interest (Gundlach et al.
2026).

Many traditional LLM benchmarks are static, or evaluated once, meaning
that they may not capture possible model changes over time and risk data
leakage (Dai et al. 2025). Performance degradation over time has been
observed in multiple studies (Roberts et al. 2023; Zhu et al. 2024; Guan
et al. 2026). The result is a call for movement toward more dynamic
benchmarks that continuously evaluate LLMs and thus avoid becoming
outdated (Chen et al. 2025). Some examples of this approach include
\texttt{Dynabench} (Kiela et al. 2021), an open-source platform for
dynamic benchmarking using human-and-model in the loop dataset creation
through a web browser and human annotators. Kiela et al. (2021) argue
that this solves the main issue of most LLMs in which performance
continuously improves on complex tasks, yet models keep failing on
simple challenge examples and falter in real-world scenarios.

Model cards---reports of newly released models published by model
providers (Mitchell et al. 2019)---are one way of understanding LLM
performance at a point in time. Example model cards from the models
included in our study are GPT-5.5 (OpenAI 2026a), Gemini 3.5 Flash
(Google DeepMind 2026), and Claude Opus 5 (Anthropic 2026b). While model
cards summarize model performance across various capabilities, they do
not account for model drift---the performance degradation of LLMs as
data and language patterns, as well as domain-specific knowledge, change
over time (Khairnar 2025). As a result, model cards become outdated as
new model versions are released and existing models are changed
post-deployment. For example, results from our first trial run of
\texttt{GardenBench}, conducted from July 20, 2026 to July 28, 2026,
suggest that the performance of GPT 5.6 Sol may have worsened in a
matter of days due to model changes post-deployment, which were reported
by OpenAI via X on July 29 (see Appendix~\ref{sec-appendix-c}).
Similarly, Chen et al. (2024) documented how the performance of GPT-3.5
and GPT 4 in math problems, degraded in a matter of months. They found
that GPT-4's accuracy in identifying prime numbers dropped from 84
percent in March 2023 to 51 percent in June 2023, suggesting that model
performance is non-static. Therefore, this literature suggests that
one-time evaluation of LLMs upon their release is insufficient, and
performance should instead be monitored over time for potential
divergence. Beyond infrequent testing, model cards are published by
model developers, reinforcing doubts about transparency and
impartiality.

\subsubsection{Reliability}\label{reliability}

In addition to running benchmarks once to calculate model performance,
it is also of interest to observe performance across repeated runs of
the same benchmark to understand model reliability. This is a critical
feature of LLM evaluation. Multiple studies examining the consistency
and reproducibility of LLM outputs have found that model accuracy varies
over time and from prompt to prompt. For example, Cui and Alexander
(2026) examined the ability of LLMs to produce the same response to the
same prompt ten-out-of-ten times and found that model output differs
even between executions with the same prompting strategy and temperature
settings. Mehta (2025) found that agent performance can drop from 60
percent on a single run to 25 percent on 8 consistent runs. This
suggests that benchmarks should be run more than once to get a full
picture of model performance. Other studies investigating model
reliability include Rabanser et al. (2026); Pinhanez et al. (2025); Ahn
and Yin (2025); Patwardhan et al. (2025).

With \texttt{GardenBench}, we introduce a new notion of reliability by
testing whether the same prompt asked to the same model at the same
time, on different days, produces the same response. We argue that the
consistency of responses across days of the week is equally an important
component of reliability. However, the expensive cost of running LLM
benchmarks does not allow for this kind of analysis, which requires
frequent recurring testing. Accordingly, \texttt{GardenBench} costs less
than \$50 USD to run the complete composition of tests, enabling us to
run the benchmark daily. This uniquely allows us to evaluate LLM
responses to the same prompt each day of the week, document day-to-day
consistency, and capture model drift.

\subsection{Related work}\label{related-work}

Public dashboards that report LLM capabilities across a variety of
skills, models, and benchmarks are already available. Some examples
include Arena, formerly ChatBot Arena (Chiang et al. 2024), and EpochAI
(Epoch AI 2026). We follow these dashboards in adopting independent and
public-facing testing of LLMs, while addressing areas for improvement.
First, Arena is a crowd-sourced approach that allows public users to
rank their preferred model responses on a variety of tasks without
disclosing the model that produced the response. This helps reduce bias
in responses, enables frequent evaluation of new models, and allows for
the input of those directly using LLMs in their daily lives. However, it
relies on the participation of the public, meaning the dashboard does
not update if people are not using it. Additionally, Arena does not
allow for quantitative comparisons between models. Similarly, EpochAI
primarily examines model accuracy against its release date without
accounting for potential trade-offs when seeking high performance, such
as the time and cost of testing. Further, models are not tested beyond
their initial release.

There are also independent web dashboards for commonly used benchmarks,
such as the \href{https://www.swebench.com/}{SWE-bench Leaderboard}
which evaluates the coding benchmark SWE-bench (Jimenez et al. 2024) on
frontier models and makes results accessible to the public. Another
example is the
\href{https://artificialanalysis.ai/evaluations/humanitys-last-exam}{Artificial
Analysis dashboard}, which displays performance on the HLE ``Humanity's
Last Exam'' benchmark, which comprises domain-specific questions
developed by field experts (Phan et al. 2026). Even though presenting
model performance on common benchmarks through live dashboards or
websites makes this information more accessible to the broader public,
these websites are not updated as frequently, possibly due to the
compounding costs of running these benchmarks frequently and as more
expensive models are released.

\texttt{GardenBench} was designed with consideration to both the
shortcomings and advantages of existing approaches to LLM benchmarking.
Namely, reliability, costs, and accessibility to the general public were
central focuses. Therefore, we propose a diversified benchmark that
includes a wide variety of everyday tasks that reflect general use
cases, many of which can be accomplished by humans. We test LLM
capabilities ranging from contextual knowledge to image recognition in a
way that is lightweight, cheap, and fast to run such that we can gauge
model performance and reliability by running the benchmark every day
over a long period. Consequently, we contribute to academic efforts in
the movement toward publicly accessible and general-use benchmarks.

\section{Data and methods}\label{sec-method}

\subsection{Developing testing suites}\label{developing-testing-suites}

In constructing our testing scheme, we follow extensive literature on
prompt engineering (Ggaliwango et al. 2024; Sahoo et al. 2025) and the
instruction-following capabilities of LLMs (Zhou et al. 2023; Wu et al.
2024). These studies find that developing specific and intentional query
structures combined with length and formatting constraints---for
instance, capitalization and punctuation---is critical in maximizing the
effectiveness of LLMs.

Our benchmark consists of 102 individual tests organized into 11
distinct testing suites. A brief overview of each suite and their
associated tasks is available in Table~\ref{tbl-suites}. We distinguish
between image- and text-based suites because some of the models we
evaluated, such as DeepSeek, were not natively capable of image
processing. Similarly, sound-based tasks were initially considered (for
example, recognizing songs from audio recordings), but ultimately
discarded as few models in 2026 natively support audio file
processing\footnote{Many large providers offer specialized models for
  audio-based tasks such as speech-to-text processing (e.g.~OpenAI's
  Whisper).}. As we evaluate model capabilities without tool use,
audio-based tasks were not feasible. In our final collection of tasks,
all correct answers could be found within the image or text in the
prompt, meaning web searching was not required. Our objective in this
approach was to evaluate raw LLM capabilities without the support of
external resources.

Tests in the same suite either evaluate LLMs on variations of the same
task or multiple closely related but distinct tasks. For instance, the
\emph{Birds} suite consists of 10 instances of the same task: ``Identify
the species of the bird shown in this image.'' Conversely, the
\emph{Books} suite contains single instances of 9 different tasks, all
of which ask the LLM to extract different information from a photograph
of a bookshelf, such as the title and author of one of the books).

Our suites represent diverse ordinary tasks that test a broad range of
LLM capabilities while being effectively cost-efficient to run. The
\emph{Ciphers}, \emph{Puzzles}, \emph{Clothing}, \emph{Transit}, and
\emph{Colours} suites test the model's ability to reason across
different domains and based on different media input types. The
\emph{Countries} and \emph{Birds} suites test base LLM knowledge, when
access to tools such as web searching was restricted. The
\emph{Haystack} and \emph{Books} suites test information retrieval. We
also included a \emph{Self-check} suite, where the LLM was prompted to
return its own model slug, a string used to identify a specific model.

We only considered tasks which could be evaluated on a binary pass or
fail. This makes the results of \texttt{GardenBench} easily
interpretable. In addition, models were only given a single chance to
respond and example responses were not provided in the prompts. This
zero-shot prompting style is intended to emulate how a human with a
quick question would interact with an LLM. Model responses were
evaluated using exact string matching between the model output and the
test's ground truth. To support reliability and scalability, the exact
same bytes were sent to each API. Moreover, prompts each demanded a
specific response structure from the LLM with no pleasantries, throat
clearing, or other excess communication. Therefore, our approach
simplifies evaluation, making results easily conceivable to broad LLM
users without sacrificing depth of analysis. An example prompt from the
\emph{Ciphers} suite is shown in Figure~\ref{fig-example-prompt}. A more
complete description of our suites is available in
Appendix~\ref{sec-appendix-a}.

\begin{table}

\caption{\label{tbl-suites}Set of suites considered in our study with
their reference names, number of tests they contain, and a one-sentence
task description.}

\centering{

\centering
\begin{tblr}[         
]                     
{                     
width={1\linewidth},
colspec={X[0.2]X[0.2]X[0.2]X[0.4]},
hline{2}={1-4}{solid, black, 0.05em},
hline{1}={1-4}{solid, black, 0.08em},
hline{13}={1-4}{solid, black, 0.08em},
column{1-4}={}{halign=l},
}                     
Suite Name & Number of Tests & Input Type & Task Description \\
Birds & 10 & Image & Identify the species of bird from its image \\
Books & 9 & Image & Identify information from an image of a bookshelf. \\
Clothes & 10 & Image & Identify information from an image of a wardrobe. \\
Colours & 12 & Image & Identify paint types and colour mixtures from an image of a paint palette. \\
Countries & 11 & Image & Identify the country from an image of its silhouette. \\
Graphs & 10 & Image & Identify the equation of the derivative from a graph of a simple function. \\
Transit & 10 & Image & Identify information from a map of a transit system. \\
Ciphers & 10 & Text & Decode a word transformed with a cipher. \\
Haystack & 8 & Text & Identify a target string within a large corpus of text. \\
Puzzles & 12 & Text & Solve logical puzzles. \\
Self-Check & 5 & Text & Answer questions about self. \\
\end{tblr}

}

\end{table}%

\begin{figure}

\centering{

\includegraphics[width=5.67in,height=\textheight,keepaspectratio]{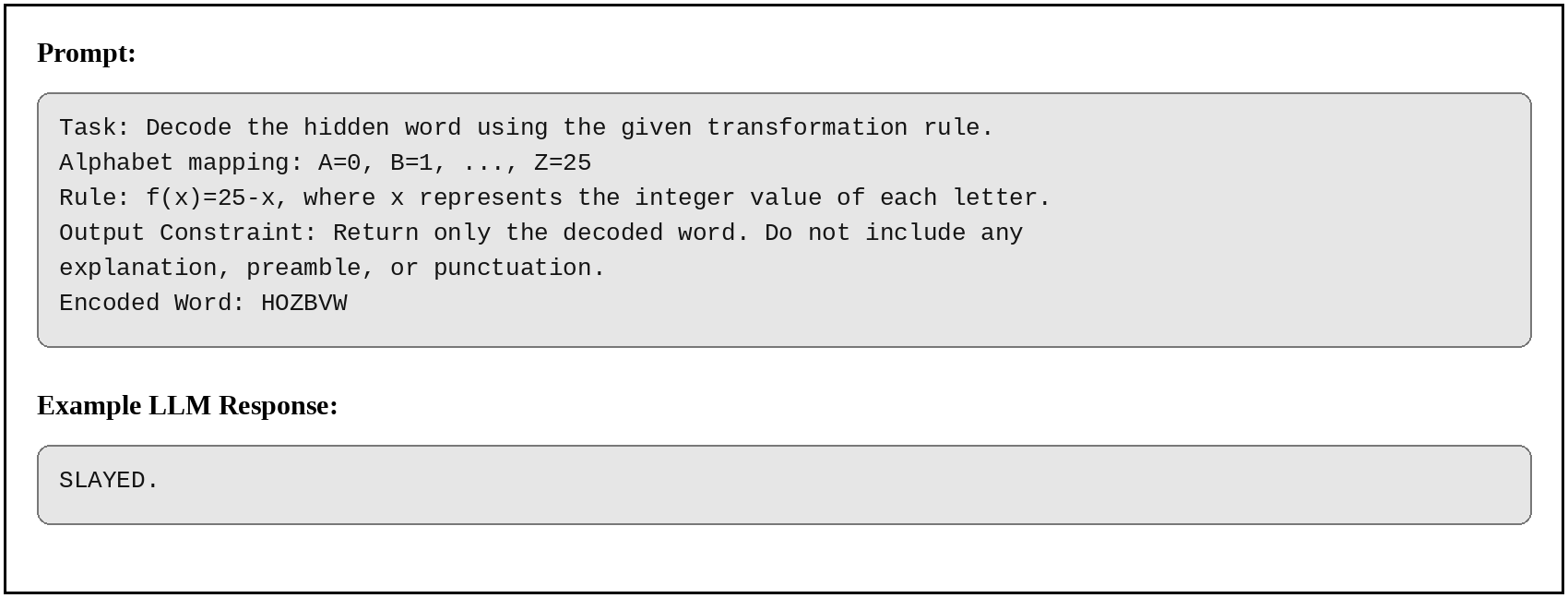}

}

\caption{\label{fig-example-prompt}Prompt used for a single instance of
the task tested by the \emph{Ciphers} suite, with an example LLM
response. The response shown is evaluated as ``Passed'' since it matches
the associated ground truth.}

\end{figure}%

\subsection{Models}\label{models}

We evaluated the performance on our benchmark of 16 different models
developed by 7 distinct providers, shown in Table~\ref{tbl-models}.
Models were prompted in Python using the OpenAI (2026b) package. The
exact API string of each model is available in
Appendix~\ref{sec-appendix-b}. This set of models includes open- and
closed-weight models developed in the U.S., Canada, and China. These
models also differ in their novelty and size. For example, Claude Haiku
4.5 was released in October 2025 and has 20 billion parameters. By
contrast, GPT 5.6 Sol was released in July 2026 and is estimated to have
2 to 4 trillion parameters. This allows us to compare model performance
across a variety of providers and versions as they develop over time.
Lastly, these models are popularly used among the general public, making
their evaluation useful to broad communities.

The set of models presented in this paper will change as providers
release new model versions and discontinue older ones. As a result,
\texttt{GardenBench} provides a permanent timeline of a model's
performance on everyday tasks. This means that changes in performance,
for instance, model drift, will be captured in the timeline of our
benchmark.

\begin{table}

\caption{\label{tbl-models}Set of models evaluated in our study,
organized by their providers, with their names, native image support
capability, and frequency of evaluation.}

\centering{

\centering
\begin{tblr}[         
]                     
{                     
width={1\linewidth},
colspec={X[0.2]X[0.2]X[0.2]X[0.4]},
hline{2}={1-4}{solid, black, 0.05em},
hline{1}={1-4}{solid, black, 0.08em},
hline{18}={1-4}{solid, black, 0.08em},
column{1-4}={}{halign=l},
}                     
Provider & Model Name & Image Support & Evaluation Frequency \\
Alibaba Cloud (via Fireworks) & Qwen 3.7 Plus & Yes & Daily \\
Anthropic & Fable 5 & Yes & Bi-weekly \\
Anthropic & Sonnet 5 & Yes & Daily \\
Anthropic & Haiku 4.5 & Yes & Daily \\
Anthropic & Opus 5 & Yes & Weekly \\
Cohere & Command-A+ & Yes & Daily \\
Cohere & Command-R+ & Yes & Daily \\
Deepseek & V4 Pro & No & Daily \\
Deepseek & V4 Flash & No & Daily \\
Google & Gemini 3.5 flash & Yes & Daily \\
Google & Gemini 3.1 pro & Yes & Weekly \\
Moonshot AI (via Fireworks) & Kimi K2.6 & Yes & Daily \\
Moonshot AI (via Fireworks) & Kimi K3 & Yes & Daily \\
OpenAI & GPT 5.6 Sol & Yes & Bi-weekly \\
OpenAI & GPT 5.4 mini & Yes & Daily \\
OpenAI & GPT 5.5 & Yes & Weekly \\
\end{tblr}

}

\end{table}%

\subsection{Evaluation metrics}\label{evaluation-metrics}

We begin by measuring aggregate performance comparatively across models
and average accuracy by suite. Analyzing aggregate performance allows
for a direct and interpretable comparison between model providers and
versions. Measuring average accuracy by suite allows us to observe the
tasks in which LLMs generally perform most successfully, indicating the
capabilities that may be most developed in LLMs.

We then analyze model accuracy as a function of several relevant
variables. First, we analyze the relationship between average dollar
cost per test and accuracy to allow for direct and transparent
comparison across models. This metric differs from price-per-token
analysis, which is often used by model providers. One limitation of this
approach is that it assumes all models output roughly the same number of
tokens to solve a given problem in order to enable comparison. However,
due to chain-of-thought reasoning, we find that some models output
significantly more tokens than others to solve the same problem. Next,
we evaluate model accuracy as a function of wall-clock time elapsed to
solve tests. Lastly, we measure model accuracy against output token use
to measure token efficiency as a separate function from cost.

Cost and time were chosen as metrics with the broad community of LLM
users in mind. For many people deciding which model to use for
non-frontier applications, cost is an important constraint and is not
often accounted for in existing benchmarks. Time is another metric that
can inform decisions of which model to use for real-world applications,
as LLMs that take seconds rather than minutes to respond may be more
effective search engines or office assistants.

\subsection{Evaluation schedule}\label{evaluation-schedule}

\begin{figure}

\centering{

\includegraphics[width=0.8\linewidth,height=\textheight,keepaspectratio]{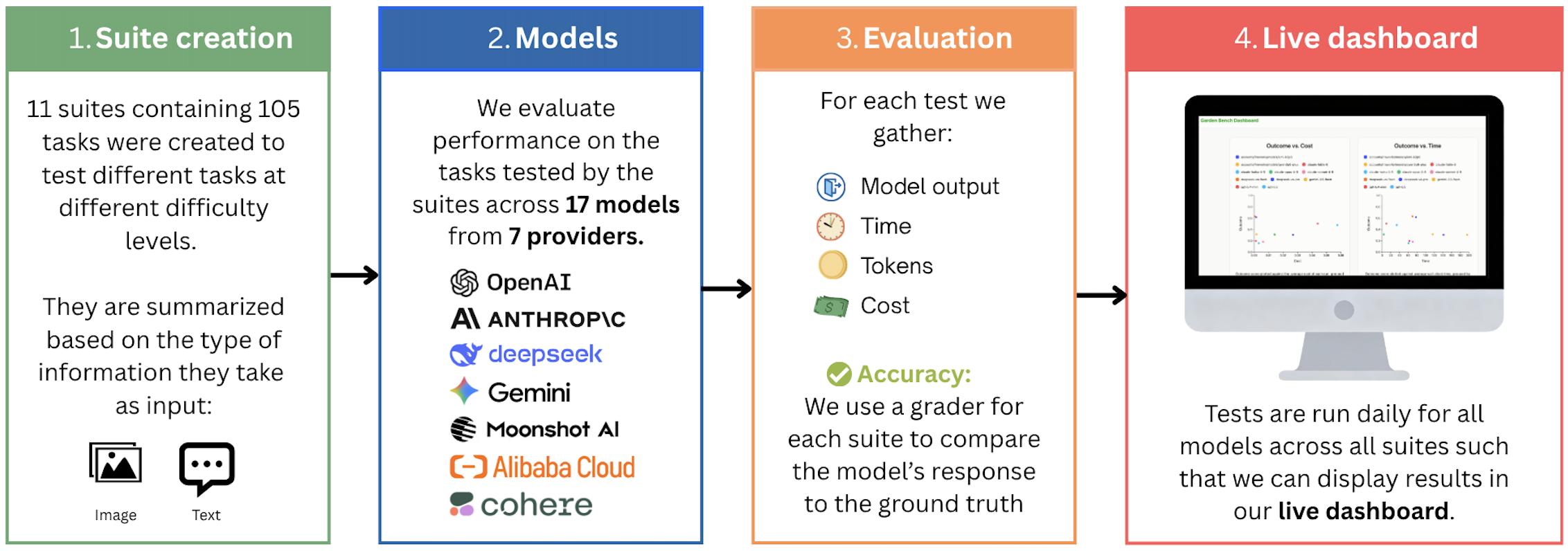}

}

\caption{\label{fig-workflow}Research workflow.}

\end{figure}

This paper analyzes results obtained from running the benchmark from
August 17, 2026 to August 31, 2026, with runs scheduled at 2:00AM
Eastern Standard Time (EST) each day. All models were evaluated daily
except for Opus 4.8, GPT 5.5, and Gemini 3.1 Pro, which were run weekly,
and Fable 5 and GPT 5.6 Sol, which were run bi-weekly due to budgetary
constraints. We use two weeks of data to demonstrate observable
fluctuations in daily performance, specifically as models frequently
progress or degrade. A cumulative dataset of historic runs was
automatically updated using the GitHub Actions workflow detailed in De
Angelis (2026). Figure~\ref{fig-workflow} presents a high-level overview
of our research workflow.

Our primary product is a publicly available live dashboard\footnote{\url{https://gardenbench.ai/}},
where visualisations of daily \texttt{GardenBench} evaluations are
updated and displayed every day. In addition to the visualizations
presented in our Results section (Section~\ref{sec-results}), the
website includes breakdowns across models and their specific
capabilities tested by the suites. This offers a more detailed
measurement than comparing aggregated accuracy across all suites.

To preserve the integrity of our experiment while running
\texttt{GardenBench} daily, we do not publish our underlying dataset.
This decision was informed by concerns about dataset contamination,
where model training data and evaluation data may intersect, enabling
LLMs to memorize responses, thus artificially enhancing performance (Xu
et al. 2024; Fu et al. 2025; Li et al. 2024). There is also less
literature on how to navigate the risk of data contamination for
emerging dynamic benchmarks (Chen et al. 2025), such as ours.

\section{Results}\label{sec-results}

\subsection{Aggregate performance}\label{aggregate-performance}

\begin{figure}

\centering{

\pandocbounded{\includegraphics[keepaspectratio]{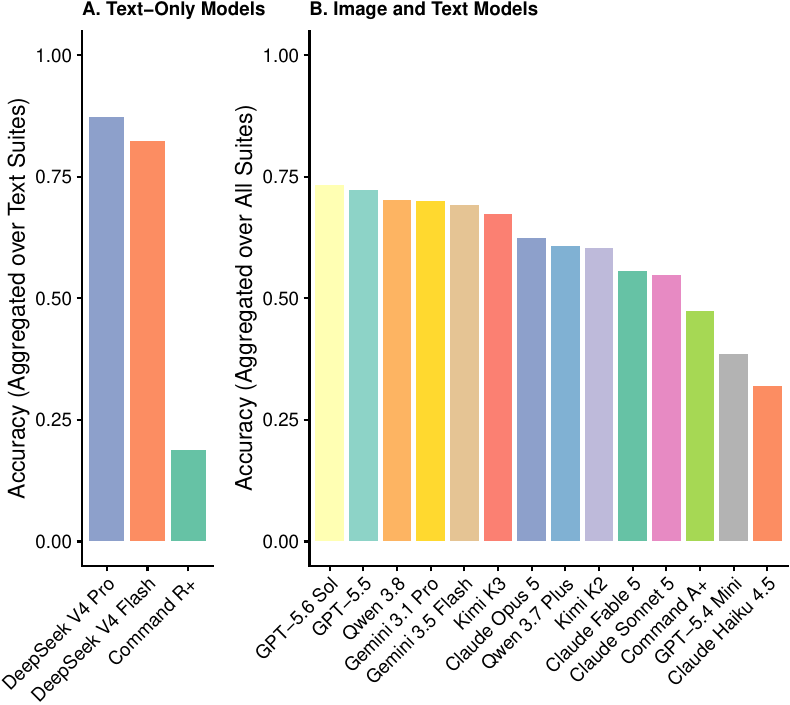}}

}

\caption{\label{fig-overall-accuracy-by-model}Performance by model,
separated into text-only models (A) and models that can handle image
input (B).}

\end{figure}

The average accuracy achieved by each model across all 11 test suites is
presented in Figure~\ref{fig-overall-accuracy-by-model}. These results
and subsequent visualizations were averaged over the period from August
17, 2026 to August 31, 2026. Replicate figures covering larger and more
recent time periods are presented on the website\footnote{\url{https://gardenbench.ai/}}.
As DeepSeek V4 Pro, DeepSeek V4 Flash, and Command R+ were not
image-capable, they were only evaluated on text-based suites (4 of the
11 total suites in Table~\ref{tbl-suites}) and are represented
separately from the others.

The highest performing model evaluated on the full benchmark was GPT-5.6
Sol, followed closely by GPT-5.5, which respectively achieved average
per-suite accuracies of 73 percent and 72 percent during our evaluation
period. Performance decreased steadily across the remaining models, with
Kimi K3, Qwen 3.8, Gemini 3.1 Pro, Gemini 3.5 Flash Claude Opus 5, Qwen
3.7 Plus, Kimi K2, Claude opus 5, Claude Fable 5, and Claude Sonnet 5
performing between 50 percent and 70 percent accuracy. Command A+,
GPT-5.4 Mini, and Claude Haiku 4.5 followed with accuracies below 50
percent.

DeepSeek V4 Pro and DeepSeek V4 Flash performed similarly on the
text-based suites, achieving average accuracies substantially higher
than Command R+---87 and 82 percent, respectively, compared to 19
percent. When other models' accuracy was evaluated on only their
text-based performance, both DeepSeek models still placed among the
strongest performers, comparable to GPT-5.5 and GPT-5.6 Sol, which
achieved average text-based accuracies of 90 percent and 87 percent,
respectively (see Appendix~\ref{sec-appendix-b}).

\subsection{Overall accuracy per
suite}\label{overall-accuracy-per-suite}

\begin{figure}

\centering{

\pandocbounded{\includegraphics[keepaspectratio]{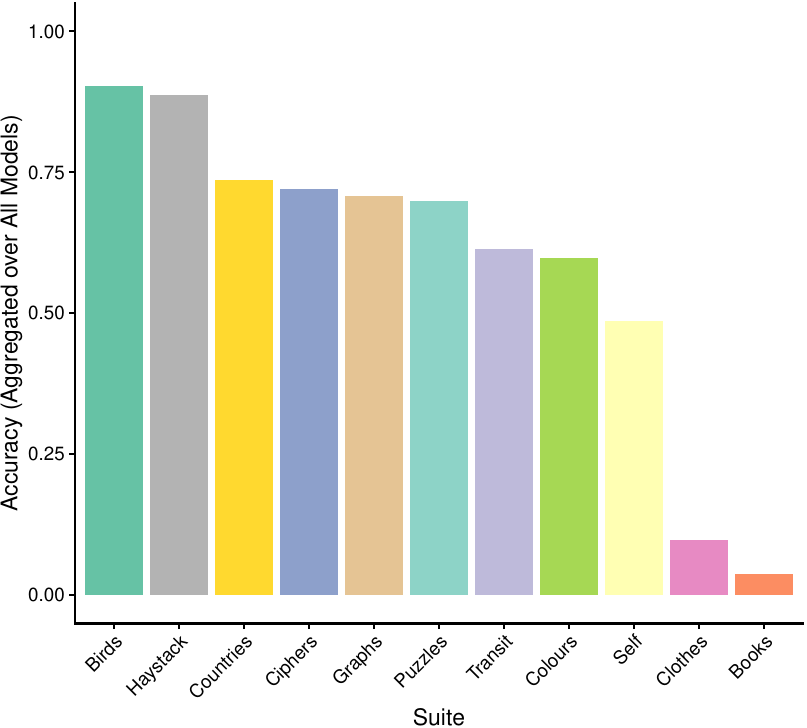}}

}

\caption{\label{fig-overall-accuracy-by-suite}Performance by suite,
averaged across all models and dates. Note that image-based suites were
evaluated on a smaller set of models, as DeepSeek V4 Flash, DeepSeek V4
Pro, and Command R+ could not process image inputs.}

\end{figure}

Next, we examine how models performed, on average, on each suite, shown
in Figure~\ref{fig-overall-accuracy-by-suite}. This demonstrates which
types of tasks and domains were generally easier or more challenging for
LLMs. Performance was highest in the \emph{Birds} and \emph{Haystack}
suites, where LLMs were asked to identify the species of bird from an
Internet-sourced image, and retrieve a target string from a block of
text respectively. Models performed similarly across the
\emph{Countries}, \emph{Ciphers}, \emph{Graphs}, \emph{Puzzles},
\emph{Transit} and \emph{Colours} suites, where average accuracy ranged
from around 50 to 60 percent. The small variation in model performance
across this combination of text- and image-based suites suggests that
models supporting both input types performed similarly whether or not
prompts included image attachments. This demonstrates generally strong
image recognition capabilities for these models. Finally, the
\emph{Clothes} and \emph{Books} suites were noticeably more challenging
than the others, as models achieved an average accuracy of less than 15
percent.

\begin{figure}

\centering{

\pandocbounded{\includegraphics[keepaspectratio]{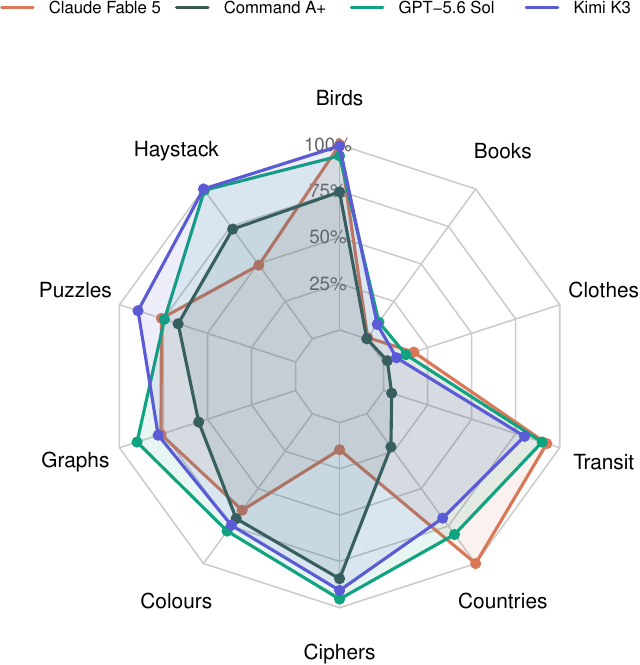}}

}

\caption{\label{fig-radar-frontier}Radar plot displaying the performance
of Claude Fable 5, Command A+, GPT-5.6 Sol, and Kimi K3 across each
suite considered. Models are colour-coded according to their model
providers.}

\end{figure}

Figure~\ref{fig-radar-frontier} displays the performance of a few select
frontier models on individual suites. We observe variation in model
performance on certain suites or groups of suites, suggesting that
aggregated accuracy metrics are not necessarily representative of a
model's specific capacities. For example, Claude Fable 5 underperformed
significantly on text-based suites relative to other frontier models,
but performed similarly on image-based suites. Model performance on
specific suites is discussed further in Appendix~\ref{sec-appendix-b}.

\subsection{Overall accuracy by run cost per
model}\label{overall-accuracy-by-run-cost-per-model}

\begin{figure}

\centering{

\pandocbounded{\includegraphics[keepaspectratio]{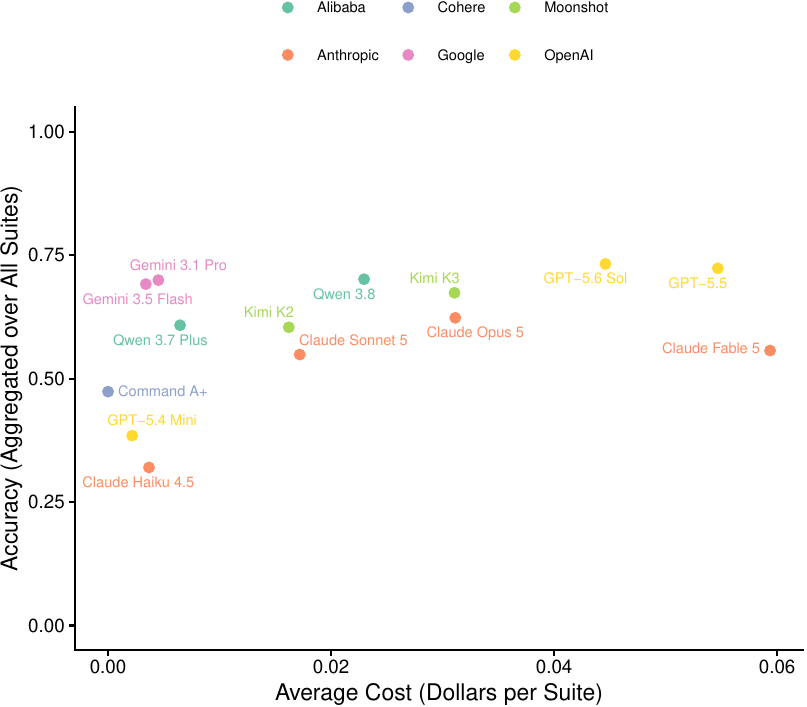}}

}

\caption{\label{fig-scatter-accuracy-cost}Relationship between accuracy
and cost averaged over all suites for every model in our study,
represented as points. Points are colour-coded according to their model
providers. DeepSeek V4 Flash, DeepSeek V4 Pro, and Command R+ were
omitted from this plot, as they were only evaluated on text-based
suites.}

\end{figure}

The complete benchmark costs approximately \$50 USD and takes up to one
hour per day to run. We plot the average per-suite accuracy of each
model against its average run cost and response time in
Figure~\ref{fig-scatter-accuracy-cost} and
Figure~\ref{fig-scatter-accuracy-time} respectively. DeepSeek V4 Pro,
DeepSeek V4 Flash, and Command R+ are not shown in these plots since
they were evaluated on a smaller subset of tests.

For around half of the models, running against the average suite cost
less than 0.02 cents to run. The most expensive models (for example,
Claude Fable 5) cost just under 0.06 cents to run. Cohere's Command A+
was the only free model considered in our study.
Figure~\ref{fig-scatter-accuracy-cost} shows an association between cost
and accuracy, suggesting that more expensive models achieve higher
accuracy than cheaper ones. One exception is the DeepSeek models, which
exhibited relatively high accuracy at marginal average costs (not
shown). This is most likely because they were evaluated against a
smaller and cheaper set of suites, although DeepSeek models did have
similar accuracies to strictly costlier models when only text-based
performance was considered.

We identify two other notable findings from this graph. First, the two
Gemini models stand out among other models, with visibly higher
accuracy-per-dollar ratios than other models. Second, while Anthropic
models have a similar cost-accuracy slope as other provider groups, they
consistently sit below other providers at every price point. Claude
Haiku 4.5 and Sonnet 4.6 cost 3 to 5 times more than GPT-5.4 Mini
without being more accurate. Claude Opus 4.8 costs more than Kimi K2 and
is 15 percent less accurate. Claude Fable 5 is the most expensive model
we tested and is 14 percent less accurate than GPT-5.5 and GPT-5.6 Sol,
which have lower costs.

\subsection{Overall accuracy by run time per
model}\label{overall-accuracy-by-run-time-per-model}

\begin{figure}

\centering{

\pandocbounded{\includegraphics[keepaspectratio]{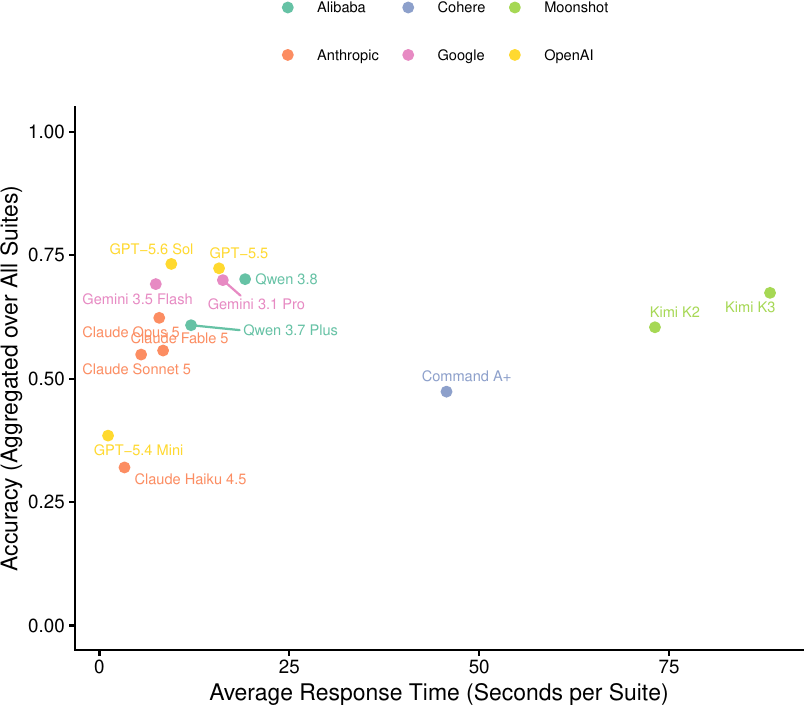}}

}

\caption{\label{fig-scatter-accuracy-time}Relationship between accuracy
and response time averaged over all suites for every model in our study,
represented as points. Points are colour-coded according to their model
providers. DeepSeek V4 Flash, DeepSeek V4 Pro, and Command R+ were
omitted from this plot, as they were only evaluated on text-based
suites.}

\end{figure}

We now turn to average response time (shown in
Figure~\ref{fig-scatter-accuracy-time}). We put a 5-minute time-out on
each test, meaning if the model took more than five minutes to complete
a task, we deemed it a failure. This decision is consistent with our
motivations for \texttt{GardenBench}, since models that take over five
minutes to respond to a simple task may not be effectively useful.

Most models took less than 25 seconds to complete suites, while Command
A+, Kimi K2, and Kimi K3 took up to 90 seconds. Kimi K2.6 timed out most
frequently, being unable to complete a task within 5 minutes 8 percent
of the time. In terms of average accuracy, Command A+ scored 47 percent,
Kimi K2.6 scored 60 percent, and Kimi K3 scored 67 percent. The
remaining models all took less than 30 seconds to respond on average,
with varying accuracy. Similarly to cost, DeepSeek models perform
relatively well given their response times, taking no more than 10
seconds to score around 85 percent accuracy on average. However, this
might again be attributed to their smaller suite subset. Prompts with
images naturally take longer for LLMs to process. Comparing average
response time on only text-based suites reveals that DeepSeek models are
taking longer to respond on average compared to similarly performant
models, such as GPT-5.5 and GPT-5.6 Sol (see
Appendix~\ref{sec-appendix-b}).

\subsection{Output token efficiency}\label{output-token-efficiency}

\begin{figure}

\centering{

\pandocbounded{\includegraphics[keepaspectratio]{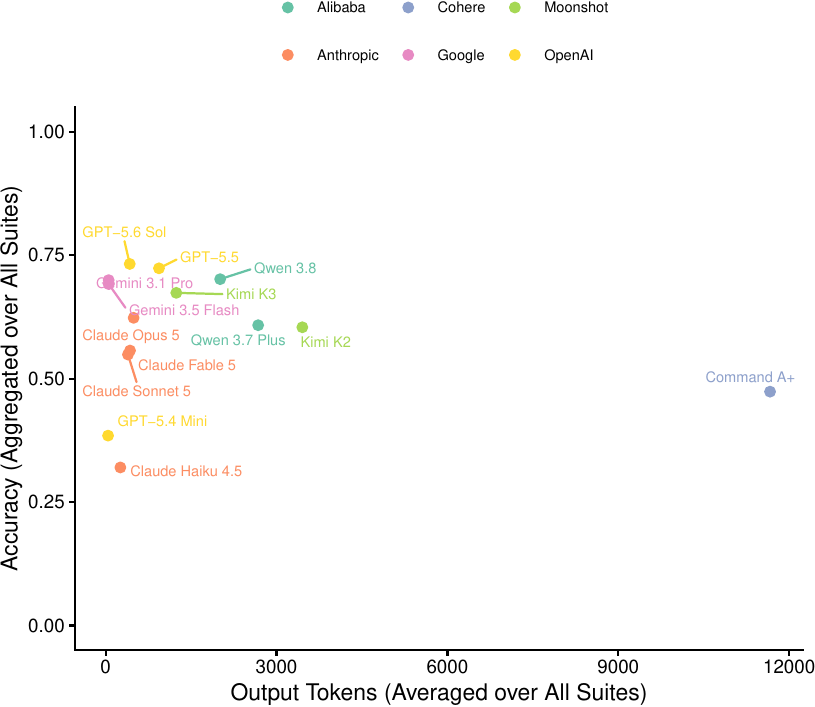}}

}

\caption{\label{fig-scatter-accuracy-output-tokens}Relationship between
accuracy and number of output tokens used averaged over all suites for
every model in our study, represented as points. Points are colour-coded
according to their model providers. DeepSeek V4 Flash, DeepSeek V4 Pro,
and Command R+ were omitted from this plot, as they were only evaluated
on text-based suites.}

\end{figure}

The average accuracy of each model by the average amount of output
tokens used is displayed in
Figure~\ref{fig-scatter-accuracy-output-tokens}. On average, Command A+
used the most output tokens by far (over 10,000), while producing an
average accuracy of 47 percent. This finding demonstrates that using
more output tokens does not necessarily result in higher accuracy.
Claude Haiku 4.5 and GPT-5.4 mini both had average per-suite accuracies
less than 50 percent, while each outputting less than 1,000 tokens.
Claude Fable 5 and Kimi K2 performed with similar accuracy, however,
Claude Fable 5 used less than 1,000 output tokens, whereas Kimi K2 used
over 3,000 output tokens.

Models from non-American providers---Canada's Cohere and China's Alibaba
and Moonshot---took more time and spent more output tokens on the
average suite than American ones. Moreover, the highest performing
models, for example, GPT-5.6 Sol, took similar amounts of time and spent
similar amounts of output tokens than some of the less successful
models, such as Claude Haiku 4.5. This suggests that differences in
model cost and size, instead of time and output token efficiency, may
better account for variation in performance from model to model.

\subsection{Accuracy over time}\label{accuracy-over-time}

\begin{figure}

\centering{

\pandocbounded{\includegraphics[keepaspectratio]{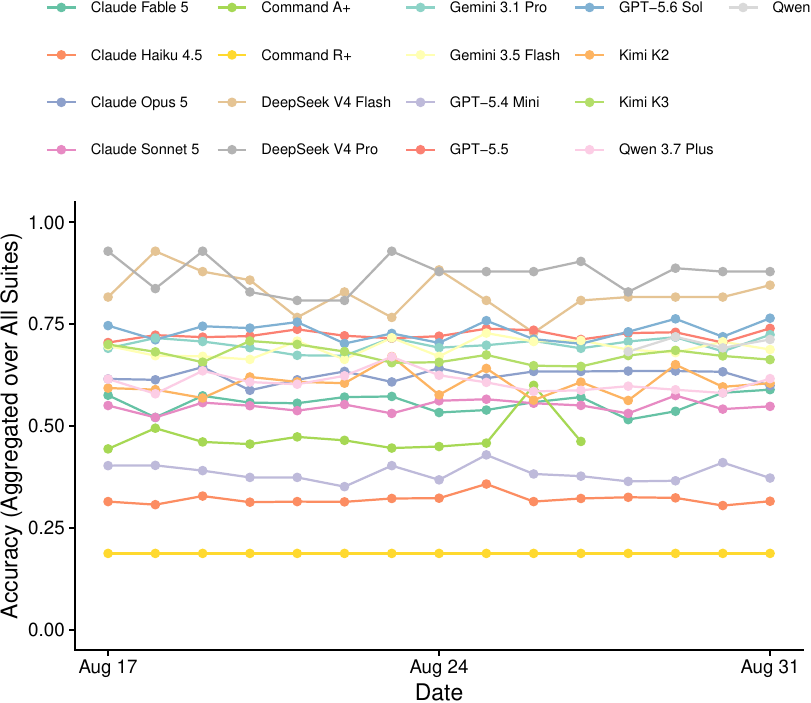}}

}

\caption{\label{fig-model-accuracy-over-time}Time series of model
performances from August 17, 2026 to August 31, 2026. Model performance
is the average accuracy scored by the model across all suites. We note
that DeepSeek V4 Flash, DeepSeek V4 Pro, and Command R+ displayed in
this Figure were only evaluated on a subset of suites.}

\end{figure}

Average daily accuracy for each model from August 17, 2026 to August 31,
2026 is displayed in Figure~\ref{fig-model-accuracy-over-time}. All
models are shown. We display results from a two-week period to capture
short-term fluctuations in performance without overcrowding the
visualization. This allows us to observe day-to-day changes in model
performance as opposed to long-term trends, which may have been more
difficult with data from a longer period.

For all models, performance appears to fluctuate slightly from day to
day. Some models like Claude Sonnet 5, Claude Opus 5, and GPT-5.4 Mini
have relatively stable performances over time, with only mild deviations
from their series average. DeepSeek V4 Flash and DeepSeek V4 Pro appear
to have the most pronounced fluctuations over time, with a wide range of
variability in scores. Others like Command A+ and Kimi K2, also have
variable performance over time, but relatively less pronounced.

These fluctuations were substantial enough to affect the daily ranking
of these models, as certain models performed better than others on some
days, and worse subsequently. For instance, on August 17, GPT-5.6 Sol
performed better than GPT-5.5. However, on August 23 and August 26,
GPT-5.5 outperformed GPT-5.6 Sol. We observe a similar case with Claude
Sonnet 5 and Claude Fable 5, as these two Anthropic models alternated in
rank from August 24 onward. This demonstrates the variability in model
performance on the same tasks over time, supporting the importance of
capturing and reporting daily evaluations. We did not observe a
consistent increasing or decreasing trend for any model. Instead, daily
fluctuations suggest that models may both progress and degrade
post-deployment.

\section{Discussion}\label{sec-discussion}

In conceptualizing \texttt{GardenBench}, we addressed two main
shortcomings in previous methods of LLM benchmarking. First,
\texttt{GardenBench} is publicly accessible, with interpretable results
posted daily to a live dashboard\footnote{\url{https://gardenbench.ai/}}.
This addresses concerns that LLM benchmarks tend to target audiences of
niche scientific communities with complex and highly specialized
evaluations (Guha et al. 2026; Koch et al. 2021). Moreover, benchmarks
are often developed and published by model providers, for example,
Anthropic (2026a)'s \texttt{BioMysteryBench}. Studies find that model
providers tend to uniquely publish benchmarks on which their models
performed well, choosing not to disclose less successful performances
(Cheng et al. 2025; Baack et al. 2026; Mitchell et al. 2019). We believe
that LLM evaluation mechanisms should be as broadly accessible as LLMs
themselves, keeping pace with the widespread use of LLMs for general,
everyday tasks by the vast public. By publishing all of our results to a
live and public dashboard, we indiscriminately show LLM capabilities in
a way that is meaningful and useful to broad LLM users.

Second, LLM benchmarks are often expensive and time-consuming, resulting
in infrequent testing (Polo et al. 2024).
Figure~\ref{fig-scatter-accuracy-cost} and
Figure~\ref{fig-scatter-accuracy-time} respectively show that, on
average, API runs for \texttt{GardenBench} cost no more than \$0.06 USD
and took no longer than 90 seconds per run (except for Kimi K2.6, which
timed out on many occasions). In total, the complete benchmark costs
approximately \$50 USD and takes up to one hour per day to run.
Systematically running our tests at the same time each day of the week
allows us to achieve a high level of reliability and scalability in our
results.

\subsection{Takeaways from model
performance}\label{takeaways-from-model-performance}

First, Figure~\ref{fig-overall-accuracy-by-suite} allowed us to observe
areas where model performance was relatively successful and others where
LLMs appeared to be underdeveloped. Specifically, string matching, image
recognition (from image-capable models), visual perception, and general
knowledge appeared to be more developed qualities of LLMs. This was
exhibited by overall performance in tasks from the \emph{Birds},
\emph{Haystack}, \emph{Countries}, \emph{Graphs}, and \emph{Colours}
suites. The tested models were slightly less successful in the
\emph{Ciphers} and \emph{Puzzles} suites, indicating that logical and
arithmetic reasoning are less developed areas. Lastly, relatively low
accuracy in the \emph{Transit}, \emph{Clothes}, and \emph{Books} suites
suggests that LLMs generally struggle with spatial reasoning questions,
contextual understanding, and information extraction.

Secondly, our results for model performance account for API run times
and costs, exhibiting trade-offs associated with achieving high
accuracy. Many high performing models took less than 25 seconds to
output responses, namely GPT-5.6 Sol and GPT-5.5, which were evaluated
on all 11 test suites. This suggests that LLM users can receive accurate
responses without waiting excessively for an answer. In the four
text-based suites, DeepSeek V4 Flash and DeepSeek V4 Pro were the
highest performing models, while also being among the most
cost-efficient (see Figure~\ref{fig-scatter-accuracy-cost}). DeepSeek V4
Flash also took relatively little time to produce responses
(approximately 10 seconds), while DeepSeek V4 Pro took almost 30
seconds. To the benefit of other LLM researchers, we demonstrate that
LLMs can be tested robustly without the expense of expensive and long
run times.

\subsection{Prompting and suite
construction}\label{prompting-and-suite-construction}

Our benchmark construction process also revealed how prompt engineering
may impact model performance. In developing testing suites, we observed
that model performance improved significantly when models were asked to
follow a specific prompt structure, for instance, in terms of
capitalization and punctuation. Asking for one-word responses,
comma-separated lists, or binary answers also allowed us to simplify our
evaluation method while preserving reliability and depth of analysis.
Anthropic models returned reasoning frequently, despite being clearly
instructed not to include additional text in their responses. On
occasions where models returned additional text, even if the answer to
the task was correct, answers were coded as incorrect. This is because
our benchmark is not only checking that the models can perform the task
correctly by providing the right answer, but also evaluating whether the
models are able to follow specific instructions by returning outputs
that followed a structured schema specified in the prompt.

Results from per-suite performance demonstrate the tasks in which
particular models struggled most (see
Figure~\ref{fig-overall-accuracy-by-suite}). Generally, models performed
with low accuracy on the \emph{Clothes} and \emph{Books} suites, which
were both based on author-collected images. The \emph{Clothes} suite,
where the task was to count the number of clothing items folded in a
closet, may have been difficult for models since some items were very
thin, dark, or looked very similar to one another. Therefore, while this
task may easily be performed by a human, the LLM was often unable to
count correctly. In the \emph{Books} suite, where models were tasked
with naming the titles and authors of books in an image of a bookshelf,
they also performed with low accuracy. On many occasions, models were
able to correctly identify the book title but failed to identify the
author. When models did not return an author name, it was given a failed
grade. It is possible that the use of external Internet resources may
have improved this result, however, this was restricted in our study.
Another suite containing author-collected images was the \emph{Colours}
suite. These were images of paintings mixed in a paint palette by one of
the authors. Even though the images were also author-collected, models
performed much better in terms of accuracy compared to the \emph{Books}
and \emph{Clothes} suite. This may be a result of specific prompt
structuring, where models were provided with a list of possible answers
and asked to select the few correct colours from the list. This proved
to significantly improve model accuracy compared to when LLMs were given
no options, as we saw in early experimentation.

Suites that involved Internet-sourced images such as \emph{Birds},
\emph{Countries} and \emph{Graphs}, performed better than suites
involving author-collected images. Although it is unclear whether the
quality of the images had an effect on the accuracy of the models in the
corresponding suites, it may have contributed to lower performance. This
may have also been due to the difficulty of the tasks themselves. The
most common error in the \emph{Birds} suite was that most models
correctly identified the Genus of the bird, but got the Species wrong.
This resulted in a completely incorrect answer, since we did not
consider partial correctness. In the \emph{Countries} and \emph{Graphs}
suites, the output from the model was entirely incorrect. For instance,
in the \emph{Countries} suite, the main error was responding with the
incorrect country. Conversely, we observed a variety of errors in the
\emph{Graphs} suite. In some cases the model was entirely incorrect,
providing the wrong equation. In other cases, the models attempted to
match the shape of the image, but failed to plug in the numbers to
actually check whether it matched, resulting in an incorrect function
with a similar shape to the ground truth.

\subsection{Daily evaluation}\label{daily-evaluation}

Lastly, the results displayed in
Figure~\ref{fig-model-accuracy-over-time}, which suggest that model
performance fluctuates daily, reinforce the importance of continuously
monitoring LLMs. We observed that some models presented more variation
in performance than others, which may impact benchmark rankings over
time. In Appendix~\ref{sec-appendix-c}, we keep results from our first
trial run of \texttt{GardenBench}, conducted from July 20 to July 28.
During this time, we observed a slight decrease in the performance of
GPT-5.6 Sol from July 26 to July 27. This was followed by an
announcement made on Twitter/X by OpenAI on July 29 of post-deployment
changes made to the model. Similar fluctuations that we see in
Figure~\ref{fig-model-accuracy-over-time} may be attributable to the
same cause. Our framework shows that it may be beneficial to
continuously evaluate models post-deployment on smaller and more
affordable benchmarks to monitor performance and make changes
accordingly. LLMs have been adopted widely nowadays as tools in both
industry and academic environments. Therefore, it is important to
supervise them as they become available to and used by the public.

\subsection{Limitations and future
directions}\label{limitations-and-future-directions}

One limitation of our benchmark is that it involves text- and
image-based prompting, even though 3 of our tested models (DeepSeek V4
Flash, DeepSeek V4 Pro, and Command-R+) were independently incapable of
image processing. It is possible that the accuracy of these models may
be overstated as they were evaluated on a reduced set of testing suites
(4 suites instead of 11). Moreover, some testing suites varied in the
number of tests included, which also could have influenced average
accuracy representations.

In addition, our tests restricted models from web searching in order to
evaluate the raw capabilities of LLMs. However, Internet information
retrieval is an important feature of LLMs that is necessary for common
general-use prompts. Therefore, future work could conduct similar
general-purpose tests with extracting knowledge from the web as an
option for LLMs. This would allow researchers to investigate whether
model performance improves when supporting resources are used, and to
what degree.

As LLM capabilities continue to progress, benchmarks should continuously
expand evaluation to the newest features of LLMs, such as image and
audio processing, as well as increasingly complex concepts and prompts.
One way to do this is by having varying difficulty levels within one
benchmark. Incorporating tasks with graduated complexity is another way
to account for modest progress. In general, we encourage future
benchmarks to continue evaluating LLM capabilities across general-use
tasks that are relevant to broad communities, and that are relatively
unaccounted for in existing benchmarks.

\section{Conclusion}\label{conclusion}

While it is important to monitor performance of models on
domain-specific tasks, it is equally important to develop benchmarks for
general aptitude. The breadth of capabilities of ML models is
increasingly as important as depth of knowledge and performance on
hyper-specific tasks. The field has moved from many specialized models
to a few models that do everything. Researchers, businesses, and
individuals now expect models that display general competence on new
tasks given only a prompt or a few examples. For many parties deciding
which model to use, the knowledge that a model is able to perform
reasonably at several office tasks is more important than the knowledge
that a model solves complex cybersecurity or mathematical problems.

This paper introduced \texttt{GardenBench}, offering an innovative
approach to measuring LLM capabilities across various general use cases
on a daily basis. We join emerging calls to reduce the gatekeeping of
LLM evaluation and help address this concern, primarily through the
publication of our live dashboard\footnote{\url{https://gardenbench.ai/}}.
The main objective of \texttt{GardenBench} is to make LLM evaluation
publicly accessible, transparent, and trustworthy. Granting broad access
to such rankings allows us to understand the most effective ways to
implement LLMs in our daily lives, as is already being done.

We also stress the importance of having benchmarks that can be run in a
lightweight manner, such that they can be run frequently to capture
model performance over time and adapted to new models effectively and
immediately once they are released.

\newpage

\appendix

\section{Benchmark Test Suites and Data Overview}\label{sec-appendix-a}

\setcounter{figure}{0}
\renewcommand{\thefigure}{A\arabic{figure}}

\setcounter{table}{0}
\renewcommand{\thetable}{A\arabic{table}}

Below we provide more detail on the suites of \texttt{GardenBench}
including a description of the tasks tested and an example of the
prompts.

\subsection{Birds}\label{birds}

The Birds test suite evaluates the capacity of Large Language Models to
process image data to perform species identification, where the model
must identify the primary calling species using its scientific name.
Performance across all experimental scenarios is automatically evaluated
via exact-match string parsing against ground-truth files, tracking
responses against \texttt{birds/01-ground\_truth.txt} for the
classification task and \texttt{birds/01-ground\_truth.txt} for the
species identification task.

\begin{figure}

\centering{

\includegraphics[width=6.2in,height=\textheight,keepaspectratio]{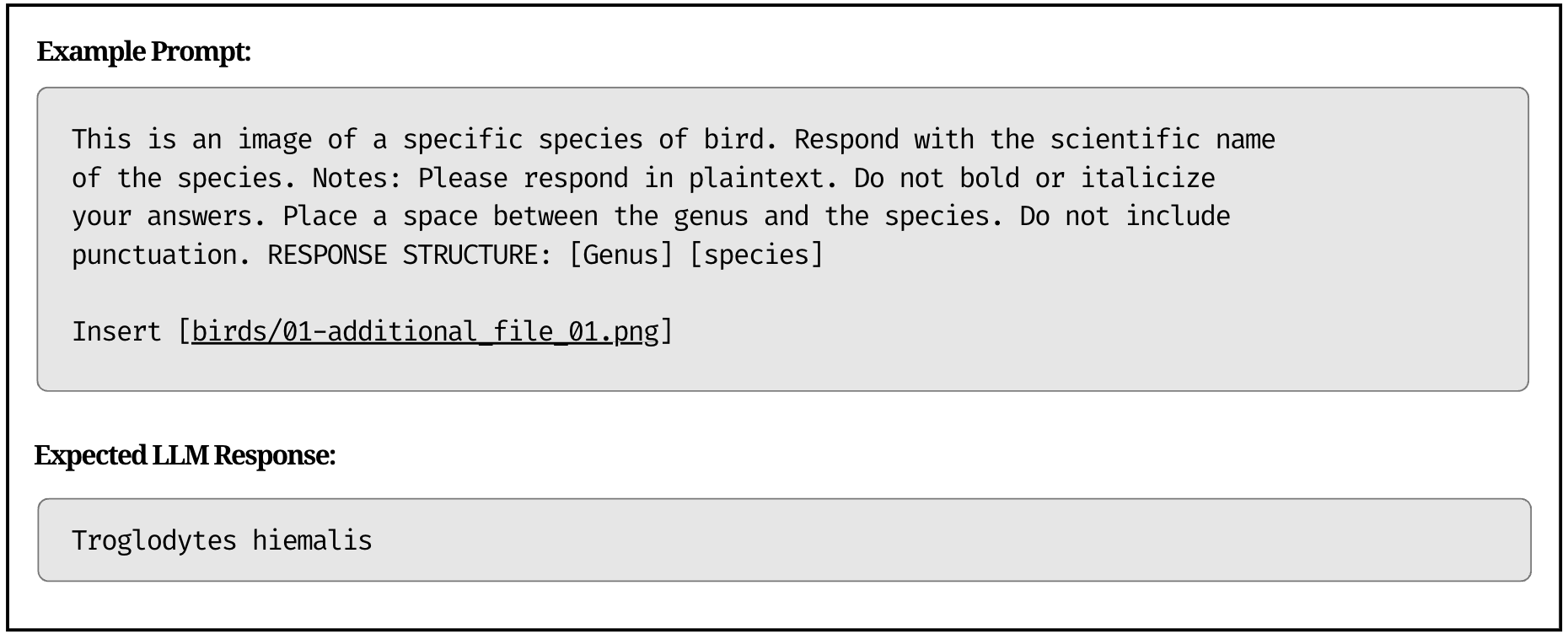}

}

\caption{\label{fig-example-prompt-birds}Prompt used for a single
instance of the task tested by the \emph{Birds} suite, with an example
LLM response. The response shown is evaluated as ``Passed'' since it
matches the associated ground truth.}

\end{figure}%

\subsection{Books}\label{books}

The Books test suite evaluates an LLM's capability to parse complex
visual data and extract structured information from images. The dataset
consists of photographs of unique bookshelves of varying sizes and book
quantities, captured directly by one of the authors with ground-truth
labels established manually. Designed to imitate tasks performed in
professional library settings, the suite features varying levels of
difficulty. It tests the model's ability to answer direct visual
questions, ranging from simple tasks like identifying the first book on
a shelf to listing out the entire bookshelf. It has also been provided a
strict formatting structure, in order to enable the comparison of its
result to a ground truth. These benchmarks require the model to parse
the image accurately and combine visual inputs with external context to
generate detailed lists of the volumes present.

For example, we have \texttt{books/01-additional\_file\_01.png} and we
ask the model the following prompts as examples.

\begin{figure}

\centering{

\includegraphics[width=6.19in,height=\textheight,keepaspectratio]{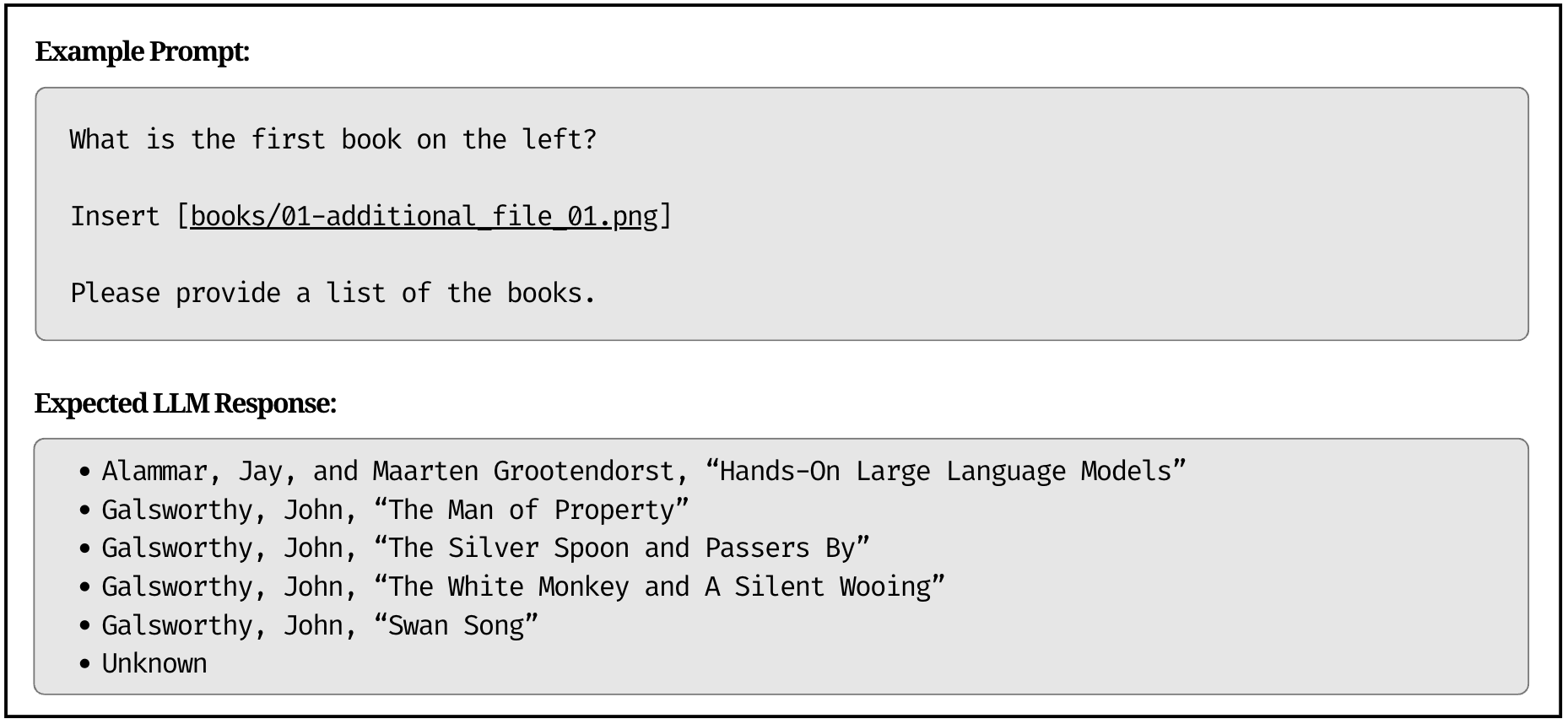}

}

\caption{\label{fig-example-prompt-books}Prompt used for a single
instance of the task tested by the \emph{Books} suite, with an example
LLM response. The response shown is evaluated as ``Passed'' since it
matches the associated ground truth.}

\end{figure}%

\subsection{Ciphers}\label{ciphers}

The Ciphers test suite evaluates an LLM's capacity for algorithmic
reasoning, character mapping adherence, and logical computation by
requiring it to decode hidden words using fixed character-level
transformation rules. The benchmark prevents models from relying on
guessing or predictive patterns by making target outputs lack semantic
or contextual cues. This could mean making a spelling error in the
target word, and evaluating whether the LLM correctly computes it or
instead incorrectly outputs the error-free word. In a baseline scenario,
models are evaluated using a zero-shot prompt with an explicit output
constraint, yielding a target string such as ``BENCHMARK''. To test the
limits of character-by-character logical execution, the suite introduces
more complex extensions, including multi-step transformations such as
f(x)=(ax+b)(mod26) and dynamic shifts based on character position.
Performance is assessed via exact-match evaluation, where the model's
output is stripped of all whitespace and compared directly against the
ground-truth target string.

\begin{figure}

\centering{

\includegraphics[width=6.19in,height=\textheight,keepaspectratio]{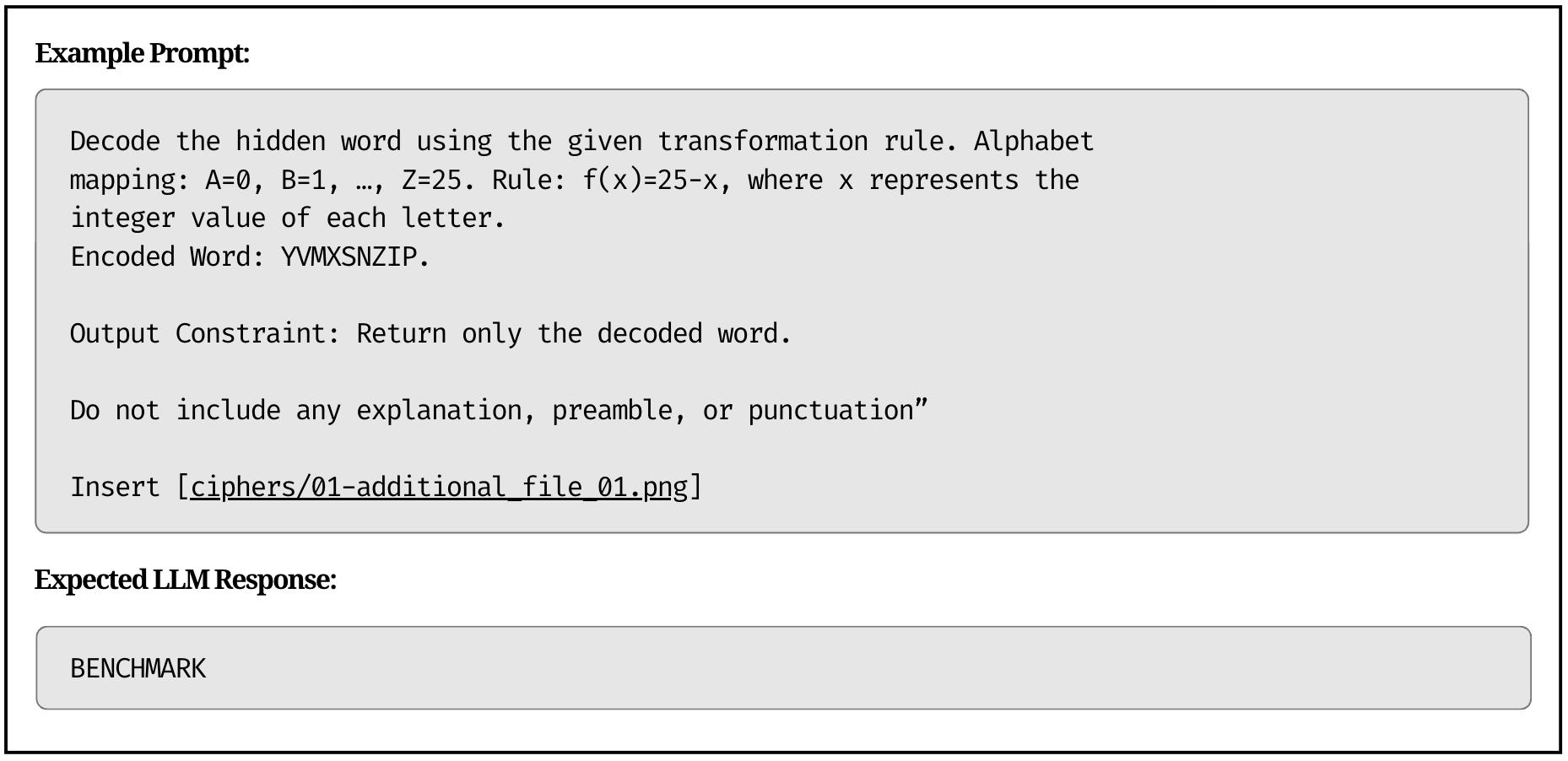}

}

\caption{\label{fig-example-prompt-ciphers}Prompt used for a single
instance of the task tested by the \emph{Ciphers} suite, with an example
LLM response. The response shown is evaluated as ``Passed'' since it
matches the associated ground truth.}

\end{figure}%

\subsection{Clothes}\label{clothes}

The Clothes test suite benchmarks an LLM's capacity for object
detection, visual counting, and classification in a singular image. The
dataset comprises ten photographs of a single closet, capturing changes
across instances, such as the removal or rearrangement of specific
items. All images were captured by one of the authors and are utilized
for the benchmark with their consent. This design evaluates the model's
image recognition counting accuracy. Evaluation is conducted using
zero-shot prompting strategies.

\begin{figure}

\centering{

\includegraphics[width=6.21in,height=\textheight,keepaspectratio]{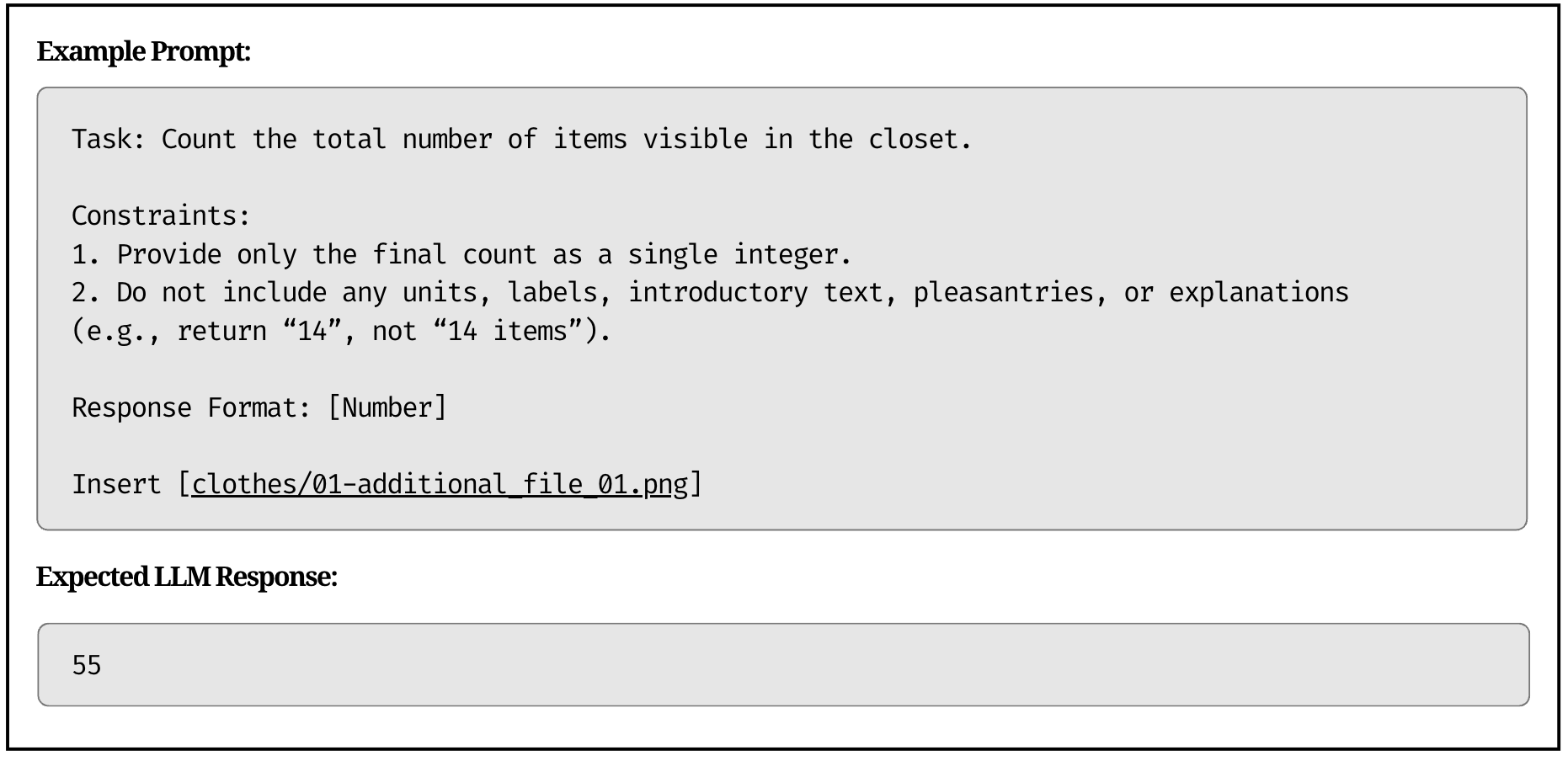}

}

\caption{\label{fig-example-prompt-clothes}Prompt used for a single
instance of the task tested by the \emph{Clothes} suite, with an example
LLM response. The response shown is evaluated as ``Passed'' since it
matches the associated ground truth.}

\end{figure}%

Model outputs are assessed via exact-match string parsing, comparing
responses directly against the corresponding verification targets in
closet/01\_ground\_truth.txt.

\subsection{Colours}\label{colours}

The Colours test suite introduces a benchmark that evaluates a model's
visual perception and reasoning. Rather than using standardized colour
codes, the dataset consists of photographs of physical paint palettes
containing mixed acrylic paints. This design assesses the model's
ability to handle artistic challenges and parse blends and textures of
colours.

The model is first prompted with a palette image and asked to identify
the kind. It is also prompted to list specific colours present. All
images were captured by one of the authors and are utilized for the
benchmark with their consent. Performance is evaluated through an
exact-match. The model's generated color list is compared directly
against a ground-truth list of the exact color names, alphabetized for
consistency.

\begin{figure}

\centering{

\includegraphics[width=6.2in,height=\textheight,keepaspectratio]{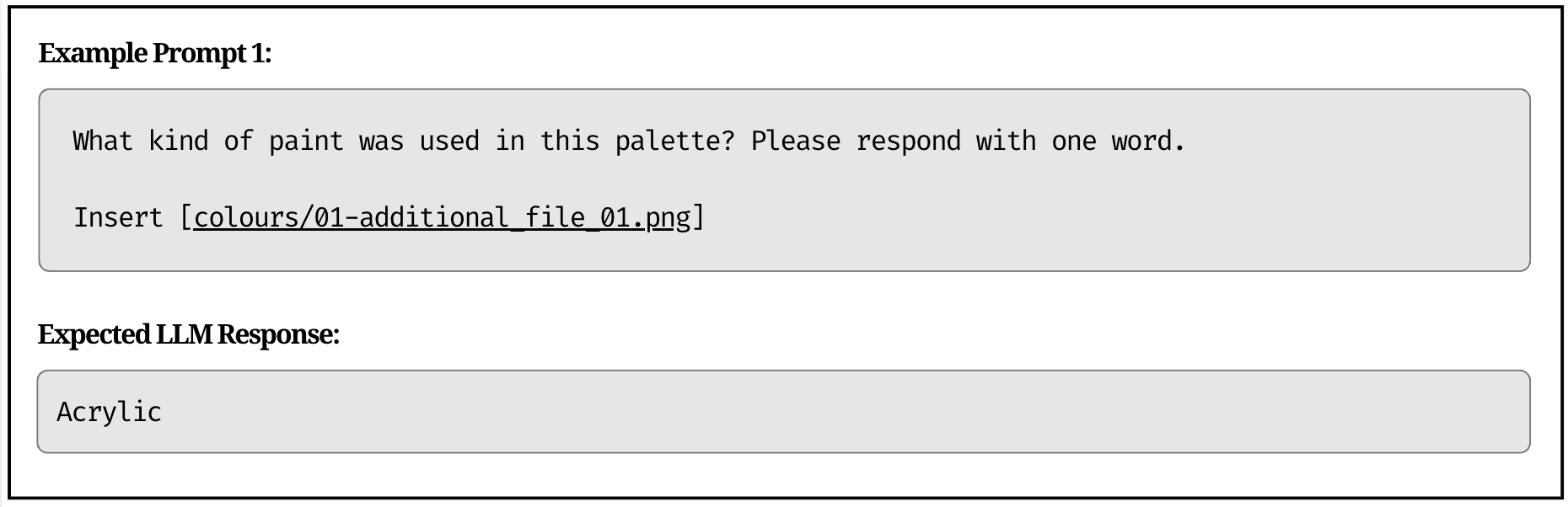}

}

\caption{\label{fig-example-prompt-colors1}Prompt used for a single
instance of the first task tested by the \emph{Colors} suite, with an
example LLM response. The response shown is evaluated as ``Passed''
since it matches the associated ground truth.}

\end{figure}%

\begin{figure}

\centering{

\includegraphics[width=6.21in,height=\textheight,keepaspectratio]{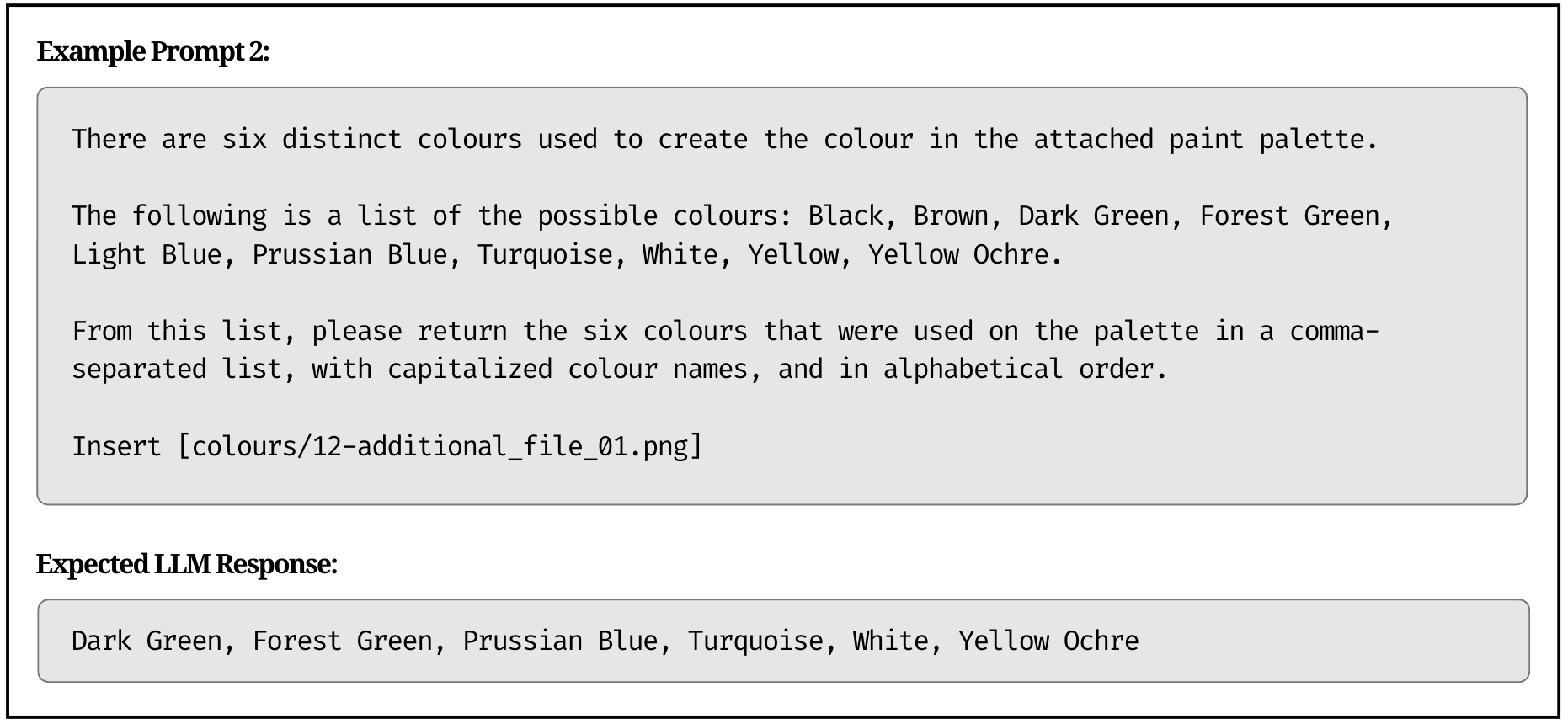}

}

\caption{\label{fig-example-prompt-colors2}Prompt used for a single
instance of the second task tested by the \emph{Colors} suite, with an
example LLM response. The response shown is evaluated as ``Passed''
since it matches the associated ground truth.}

\end{figure}%

\subsection{Countries}\label{countries}

The Countries test suite evaluates an LLM's visual pattern recognition
and geographic knowledge retrieval by challenging it to identify a
specific country based on the outline of its geographic boundary.
Inspired by the popular geography game
\href{https://worldle.teuteuf.fr}{Worldle}, this benchmark forces models
to process spatial contours and map those shapes to geographic data
without relying on text labels, regional context, or surrounding
borders. All images of country outlines were sourced from
\href{https://www.canva.com/}{Canva} under the Free Plan, and provided
to the model as .png files. Evaluation is conducted across zero-shot and
one-shot scenarios. In the zero-shot setting, the model is provided with
a geographic file and a direct identification directive.

\begin{figure}

\centering{

\includegraphics[width=6.2in,height=\textheight,keepaspectratio]{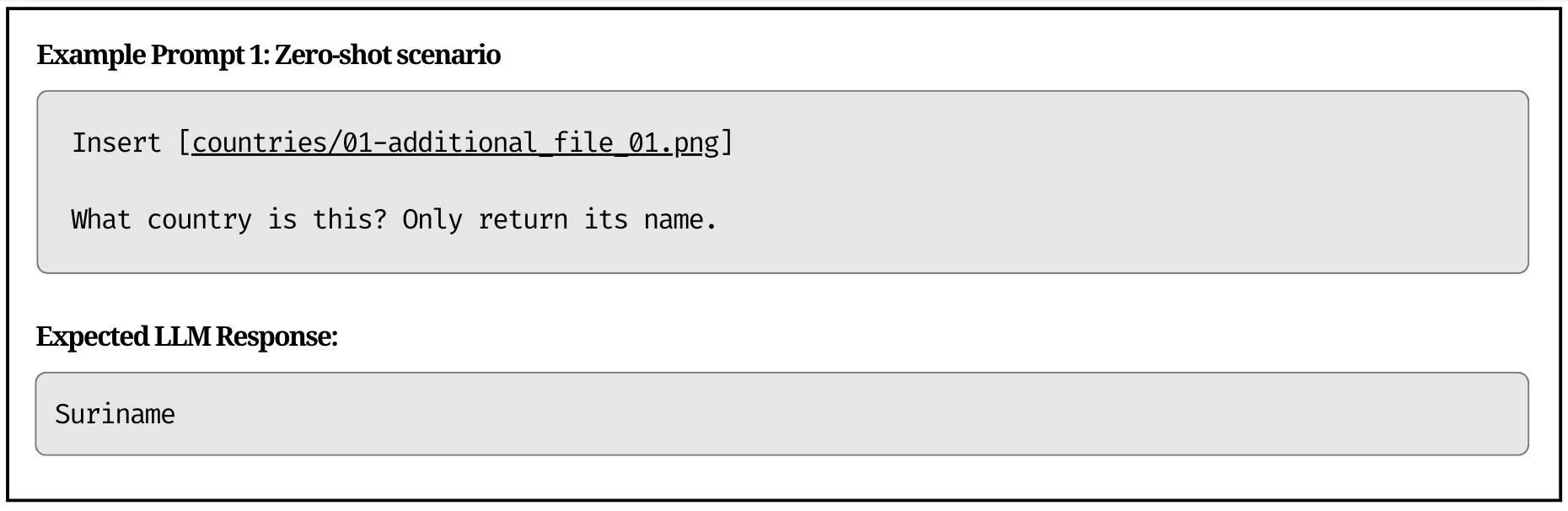}

}

\caption{\label{fig-example-prompt-countries1}Prompt used for a single
instance of the zero-shot scenario task tested by the \emph{Countries}
suite, with an example LLM response. The response shown is evaluated as
``Passed'' since it matches the associated ground truth.}

\end{figure}%

\begin{figure}

\centering{

\includegraphics[width=6.19in,height=\textheight,keepaspectratio]{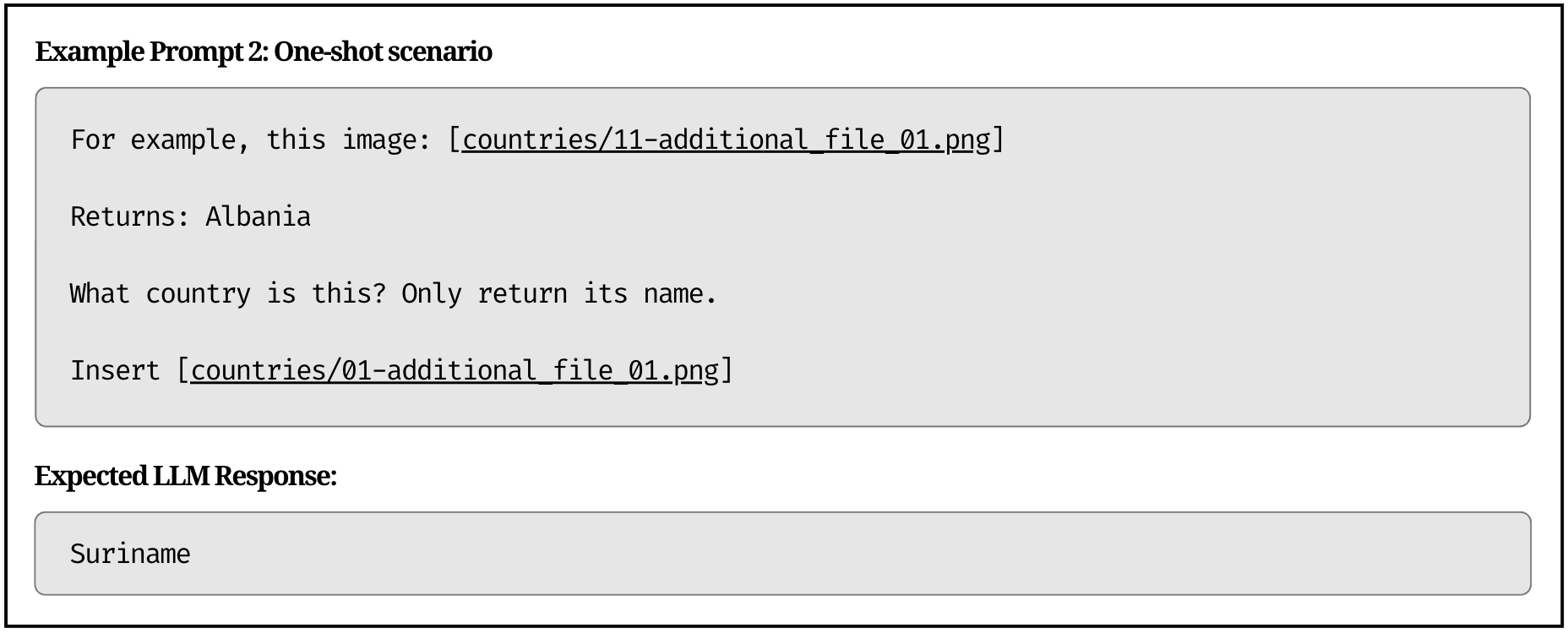}

}

\caption{\label{fig-example-prompt-countries2}Prompt used for a single
instance of the task tested by the \emph{Countries} suite, with an
example LLM response. The response shown is evaluated as ``Passed''
since it matches the associated ground truth.}

\end{figure}%

\subsection{Graphs}\label{graphs}

The Graphs test suite measures geometric reasoning, visual data
interpretation, and image-generation capabilities by prompting models to
perform math operations on plots. Specifically, the benchmark requires
the model to interpret the visual curve of a function and generate the
corresponding graph of its first derivative, or conversely, to get the
original function from a provided derivative graph. This task tests the
model's capacity to visually identify mathematical knowledge like
inflection points, local extrema, and asymptotic behavior, and translate
these into an accurate representation. The suite uses zero-shot
prompting, where the configuration asks for the mathematical graph once
provided the image input. Evaluation will then involve testing the LLM
as a judge, and determine through a Yes/No if the images are identical.

\begin{figure}

\centering{

\includegraphics[width=6.21in,height=\textheight,keepaspectratio]{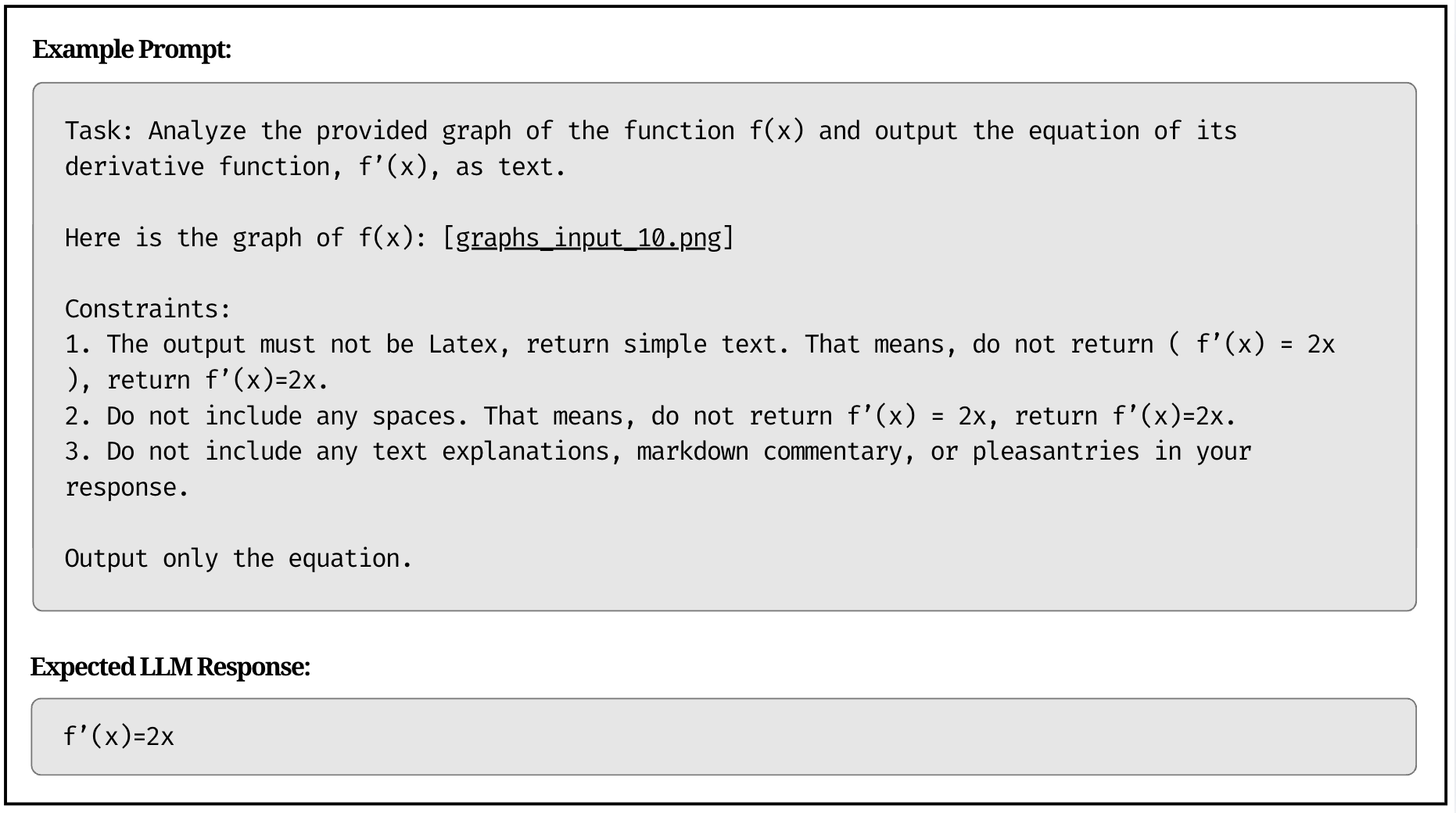}

}

\caption{\label{fig-example-prompt-graphs}Prompt used for a single
instance of the task tested by the \emph{Graphis} suite, with an example
LLM response. The response shown is evaluated as ``Passed'' since it
matches the associated ground truth.}

\end{figure}%

\subsection{Puzzles}\label{puzzles}

The Puzzles test suite benchmarks an LLM's capacity for context-window
rule adherence, logical reasoning, and multi-step deduction by
evaluating its performance on Knights and Knaves logic puzzles (Smullyan
1981). In these scenarios, a set of characters are designated as either
knights (who strictly speak the truth) or knaves (who strictly lie) and
make self-referential statements about one another's identities. These
puzzles are deterministic and solvable via formal inference engines. We
measure how reliably LLMs deduce pure logic compared to standard
computation. The dataset comprises 10 structurally distinct puzzles
featuring scaled complexities ranging from 3 to 6 characters and diverse
statement types. The puzzles and corresponding solutions were
programmatically generated and verified by adapting the open-source
TruthQuest framework (Mondorf and Plank 2024).

\begin{figure}

\centering{

\includegraphics[width=6.19in,height=\textheight,keepaspectratio]{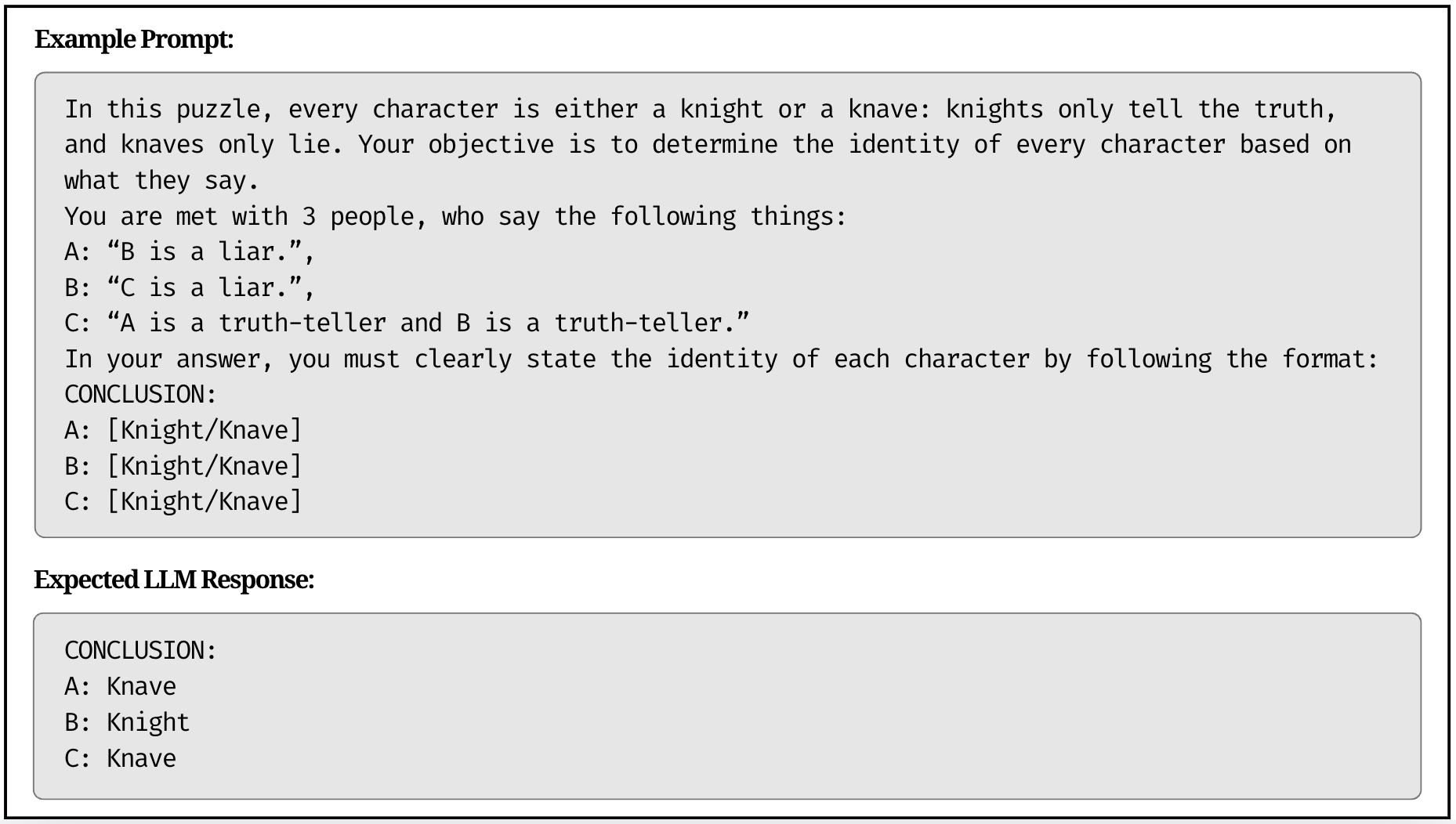}

}

\caption{\label{fig-example-prompt-puzzles}Prompt used for a single
instance of the task tested by the \emph{Puzzles} suite, with an example
LLM response. The response shown is evaluated as ``Passed'' since it
matches the associated ground truth.}

\end{figure}%

\subsection{Transit}\label{transit}

\begin{figure}

\centering{

\includegraphics[width=6.19in,height=\textheight,keepaspectratio]{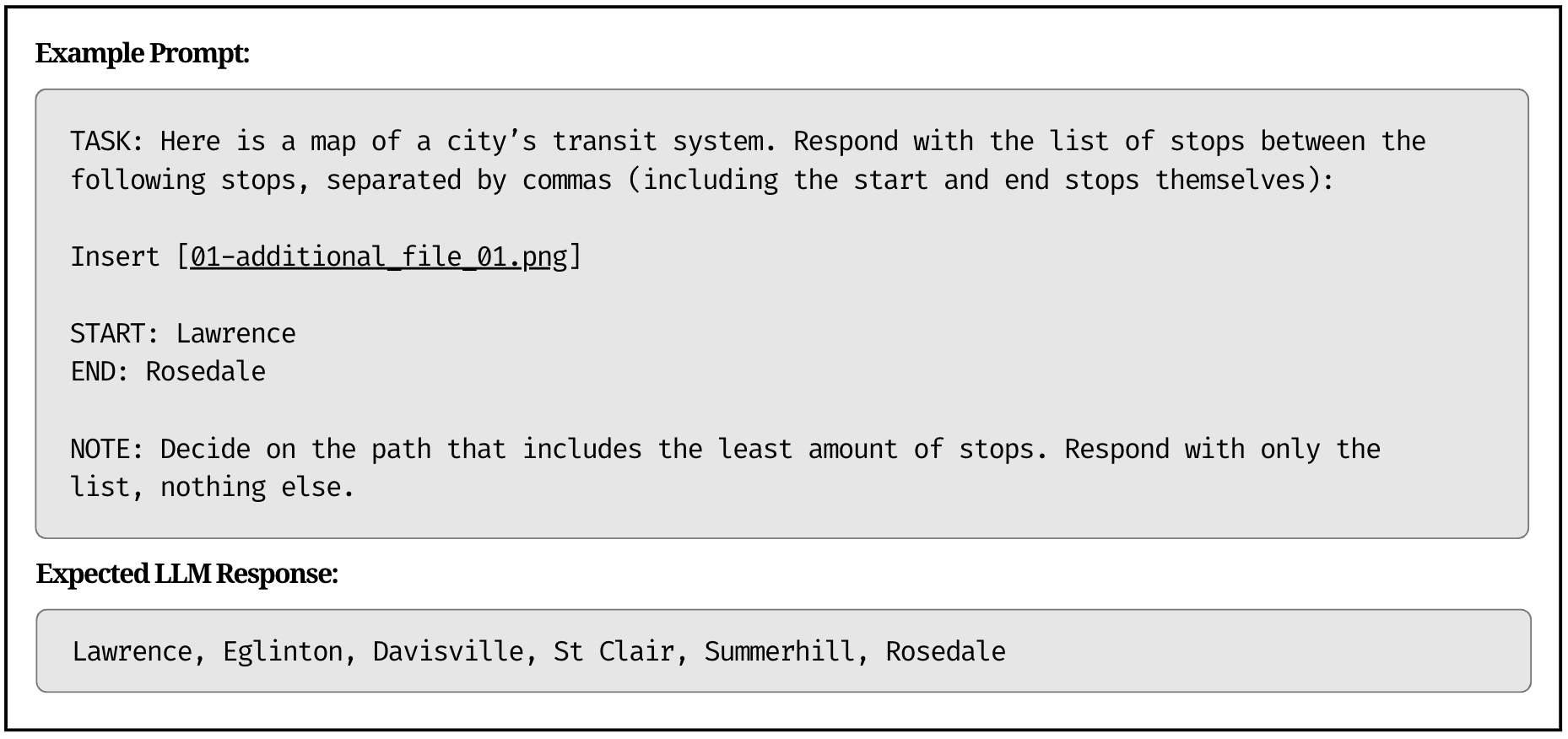}

}

\caption{\label{fig-example-prompt-transit}Prompt used for a single
instance of the task tested by the \emph{Transit} suite, with an example
LLM response. The response shown is evaluated as ``Passed'' since it
matches the associated ground truth.}

\end{figure}%

\subsection{Ascii Haystack and
Haystack}\label{ascii-haystack-and-haystack}

The Haystack Suite tests whether the model can identify a target string
from within a large corpus of text. The main difference between Ascii
Haystack and Haystack is that Ascii Haystack embeds the relevant
information within random letters and symbols, whereas Haystack embeds
random letters and symbols within coherent prose.

\begin{figure}

\centering{

\includegraphics[width=6.19in,height=\textheight,keepaspectratio]{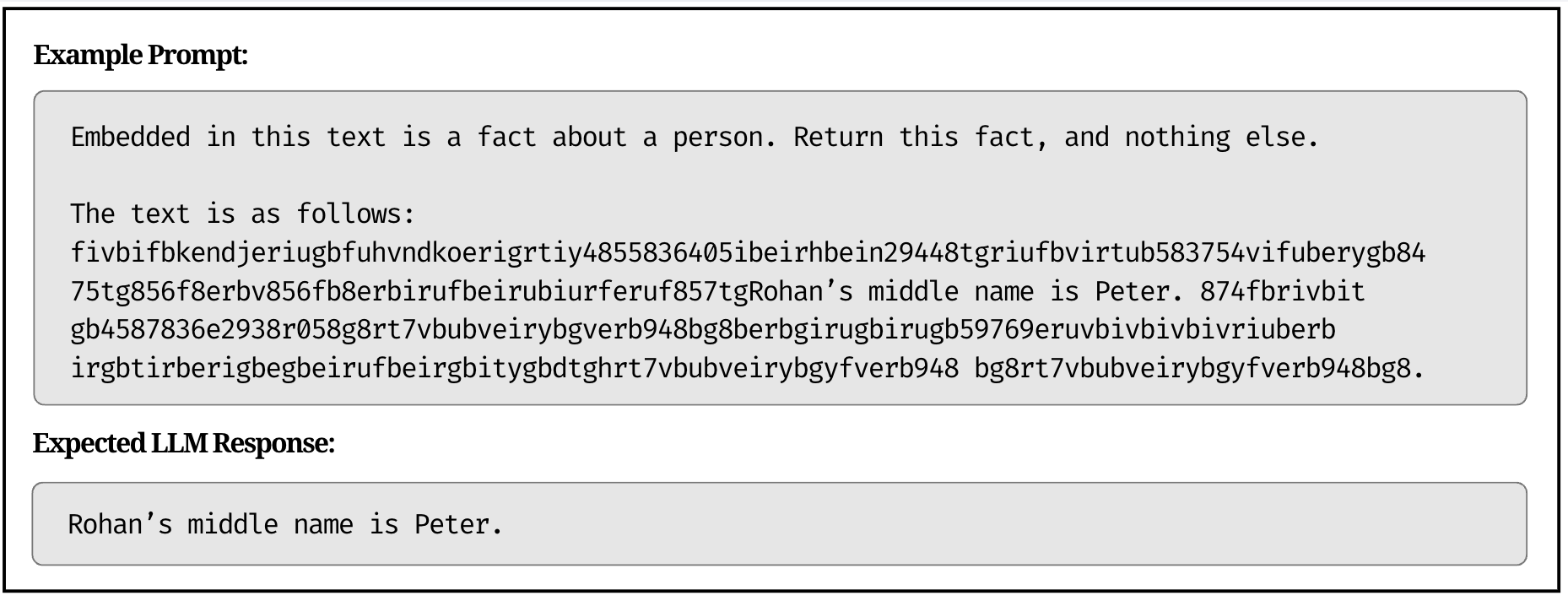}

}

\caption{\label{fig-example-prompt-haystack1}Prompt used for a single
instance of the task tested by the \emph{Haystack} suite, in the Ascii
Haystack scenario, with an example LLM response. The response shown is
evaluated as ``Passed'' since it matches the associated ground truth.}

\end{figure}%

\begin{figure}

\centering{

\includegraphics[width=6.19in,height=\textheight,keepaspectratio]{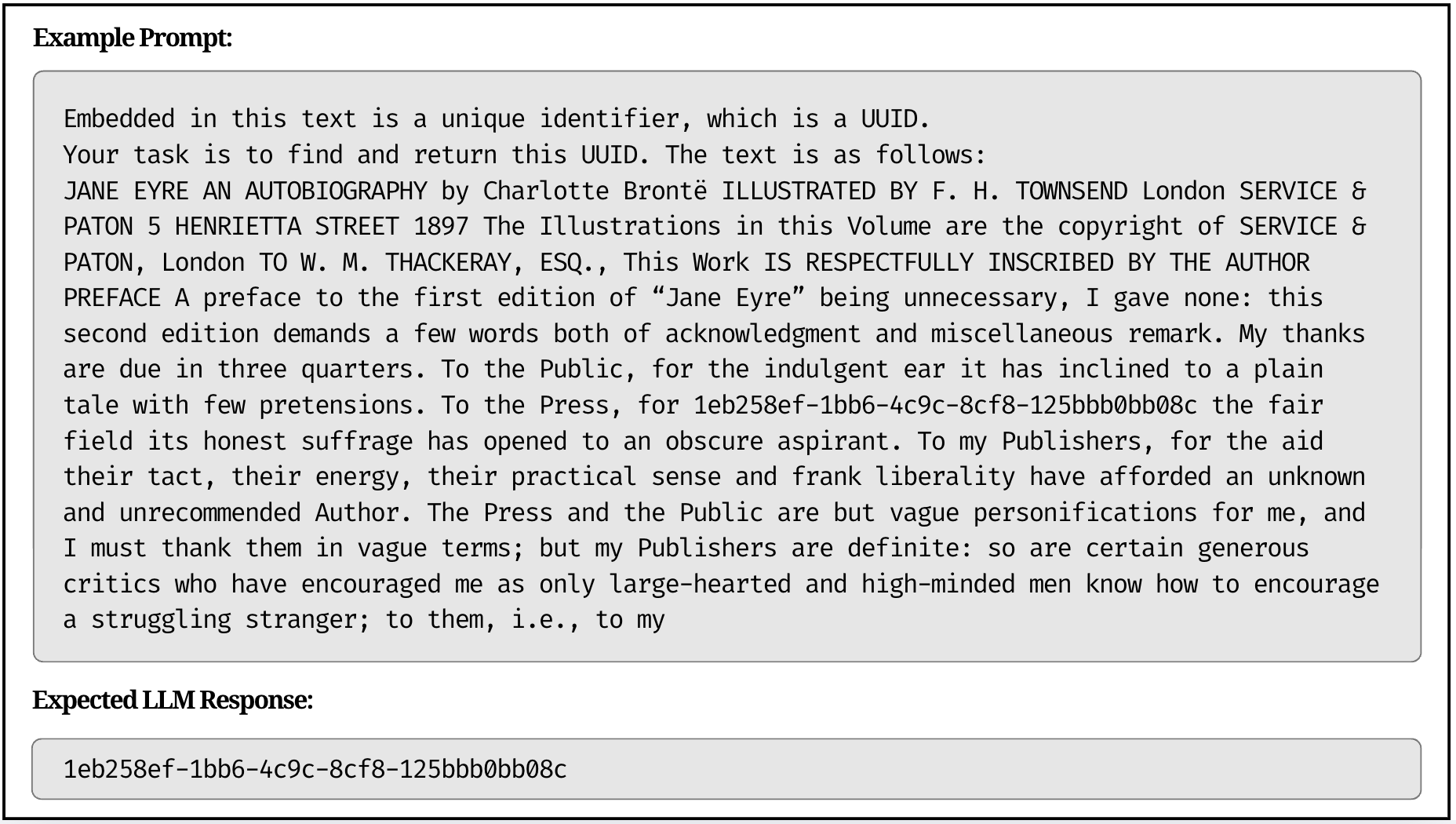}

}

\caption{\label{fig-example-prompt-haystack2}Prompt used for a single
instance of the task tested by the \emph{Haystack} suite, in the regular
Haystack scenario, with an example LLM response. The response shown is
evaluated as ``Passed'' since it matches the associated ground truth.}

\end{figure}%

\subsection{Self Check}\label{self-check}

The Self Check test suite evaluates the model's ability to answer
questions about the environment in which it is operating, for instance,
``What date is it today?''. This test encompasses a variety of text
prompts and tests the model's real time knowledge retrieval and system
awareness.

\begin{figure}

\centering{

\includegraphics[width=6.21in,height=\textheight,keepaspectratio]{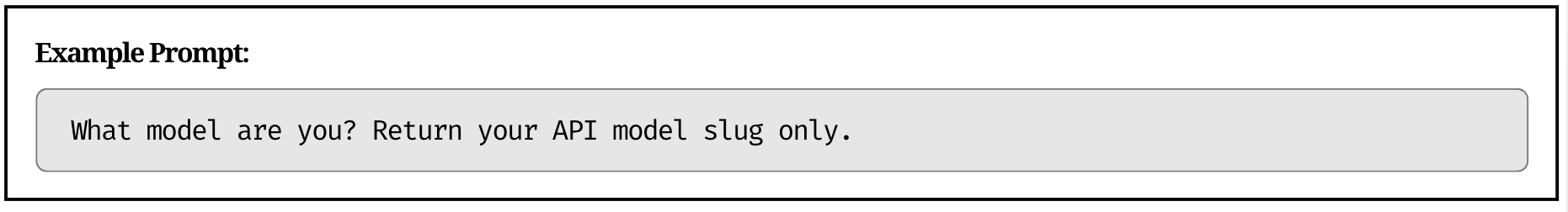}

}

\caption{\label{fig-example-prompt-selfcheck}Prompt used for a single
instance of the task tested by the \emph{Self Check} suite, with an
example LLM response. The response shown is evaluated as ``Passed''
since it matches the associated ground truth.}

\end{figure}%

\section{Additional tables and graphs}\label{sec-appendix-b}

\setcounter{figure}{0}
\renewcommand{\thefigure}{A\arabic{figure}}

\setcounter{table}{0}
\renewcommand{\thetable}{A\arabic{table}}

\begin{table}

\caption{\label{tbl-accuracy-by-suite-by-model}Full table of accuracies
by models and suites. Accuracies are averaged across time. We note that
DeepSeek V4 Flash, DeepSeek V4 Pro, and Command R+ were only evaluated
on text-based suites.}

\centering{

\centering
\begin{tblr}[         
]                     
{                     
colspec={Q[]Q[]Q[]Q[]Q[]Q[]Q[]Q[]Q[]Q[]Q[]Q[]},
hline{2}={3-12}{solid, black, 0.03em},
hline{2}={2}{solid, black, 0.03em, l=-0.5},
hline{3}={1-12}{solid, black, 0.05em},
hline{1}={1-12}{solid, black, 0.08em},
hline{20}={1-12}{solid, black, 0.08em},
row{2}={}{font=\bfseries},
cell{1}{1}={}{font=\bfseries, halign=c},
cell{1}{10}={}{halign=c},
cell{1}{11}={}{halign=c},
cell{1}{12}={}{halign=c},
cell{1}{2}={c=11}{halign=c},
cell{1}{3}={}{halign=c},
cell{1}{4}={}{halign=c},
cell{1}{5}={}{halign=c},
cell{1}{6}={}{halign=c},
cell{1}{7}={}{halign=c},
cell{1}{8}={}{halign=c},
cell{1}{9}={}{halign=c},
cell{3-19}{1}={}{font=\bfseries},
}                     
& Suite &  &  &  &  &  &  &  &  &  &  \\
Model & Birds & Books & Ciphers & Clothes & Colours & Countries & Puzzles & Graphs & Haystack & Transit & Self \\
Claude Fable 5 & 1 & 0.0079 & 0.15 & 0.1714 & 0.644 & 1 & 0.76 & 0.76 & 0.49 & 0.927 & 0.2 \\
Claude Haiku 4.5 & 0.8 & 0 & 0.02 & 0.08 & 0.444 & 0.55 & 0 & 0.24 & 0.99 & 0 & 0.4 \\
Claude Opus 5 & 1 & 0.0296 & 0.85 & 0.1689 & 0.052 & 1 & 0.91 & 0.79 & 0.83 & 0.807 & 0.41 \\
Claude Sonnet 5 & 0.93 & 0 & 0.73 & 0.0067 & 0.567 & 1 & 0.82 & 0.64 & 0.58 & 0.353 & 0.4 \\
Command A+ & 0.75 & 0 & 0.84 & 0.0211 & 0.7 & 0.22 & 0.67 & 0.55 & 0.73 & 0.046 & 0.57 \\
Command R+ & NA & NA & 0 & NA & NA & NA & 0 & NA & 0.75 & NA & 0 \\
DeepSeek V4 Flash & NA & NA & 0.99 & NA & NA & NA & 0.77 & NA & 1 & NA & 0.53 \\
DeepSeek V4 Pro & NA & NA & 0.99 & NA & NA & NA & 0.88 & NA & 1 & NA & 0.61 \\
GPT-5.4 Mini & 0.77 & 0 & 0.11 & 0.06 & 0.589 & 0.22 & 0.12 & 0.55 & 0.94 & 0.08 & 0.8 \\
GPT-5.5 & 0.9 & 0.0963 & 1 & 0.0267 & 0.717 & 0.75 & 0.79 & 0.89 & 1 & 0.98 & 0.8 \\
GPT-5.6 Sol & 0.94 & 0.1111 & 0.95 & 0.1267 & 0.783 & 0.81 & 0.74 & 0.9 & 0.99 & 0.9 & 0.8 \\
Gemini 3.1 Pro & 0.9 & 0.0148 & 0.99 & 0.0933 & 0.611 & 1 & 0.92 & 0.76 & 1 & 0.86 & 0.55 \\
Gemini 3.5 Flash & 0.94 & 0.0296 & 0.95 & 0.1267 & 0.661 & 1 & 0.91 & 0.73 & 1 & 0.859 & 0.4 \\
Kimi K2 & 0.92 & 0.037 & 0.93 & 0.08 & 0.683 & 0.95 & 0.88 & 0.63 & 0.85 & 0.46 & 0.21 \\
Kimi K3 & 0.99 & 0.0963 & 0.91 & 0.0733 & 0.744 & 0.7 & 0.89 & 0.78 & 1 & 0.8 & 0.43 \\
Qwen 3.7 Plus & 0.9 & 0.0667 & 0.85 & 0.1689 & 0.5 & 0.39 & 0.89 & 0.78 & 0.93 & 0.635 & 0.59 \\
Qwen 3.8 & 0.9 & 0.0278 & 0.97 & 0.175 & 0.667 & 0.73 & 0.9 & 0.9 & 1 & 0.9 & 0.55 \\
\end{tblr}

}

\end{table}%

\begin{table}

\caption{\label{tbl-model-api-strings}Full table of exact model API
strings.}

\centering{

\centering
\begin{tblr}[         
]                     
{                     
width={0.8\linewidth},
colspec={X[0.2]X[0.2]X[0.4]},
hline{2}={1-3}{solid, black, 0.05em},
hline{1}={1-3}{solid, black, 0.08em},
hline{18}={1-3}{solid, black, 0.08em},
column{1-3}={}{halign=l},
}                     
Provider & Model Name & API String \\
Alibaba Cloud (via Fireworks) & Qwen 3.7 Plus & accounts/fireworks/models/qwen3p7-plus \\
Anthropic & Fable 5 & claude-fable-5 \\
Anthropic & Haiku 4.5 & claude-haiku-4-5 \\
Anthropic & Opus 5 & claude-opus-5 \\
Anthropic & Sonnet 5 & claude-sonnet-5 \\
Cohere & Command-A+ & command-a-plus-05-2026 \\
Cohere & Command-R+ & command-r-plus-08-2024 \\
Deepseek & V4 Flash & deepseek-v4-flash \\
Deepseek & V4 Pro & deepseek-v4-pro \\
Google & Gemini 3.1 pro & gemini-3.1-pro-preview \\
Google & Gemini 3.5 flash & gemini-3.5-flash \\
Moonshot AI (via Fireworks) & Kimi K2.6 & accounts/fireworks/models/kimi-k2p6 \\
Moonshot AI (via Fireworks) & Kimi K3 & kimi-k3 \\
OpenAI & GPT 5.4 mini & gpt-5.4-mini \\
OpenAI & GPT 5.5 & gpt-5.5 \\
OpenAI & GPT 5.6 Sol & gpt-5.6-sol \\
\end{tblr}

}

\end{table}%

\begin{figure}

\centering{

\pandocbounded{\includegraphics[keepaspectratio]{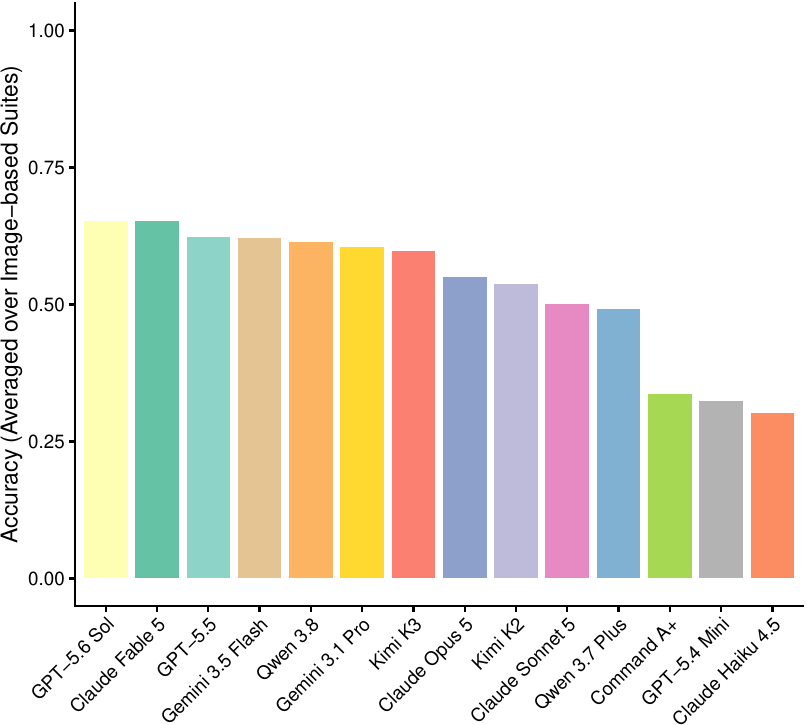}}

}

\caption{\label{fig-overall-accuracy-by-model-image}Performance by
model, on only image-based suites.}

\end{figure}%

\begin{figure}

\centering{

\pandocbounded{\includegraphics[keepaspectratio]{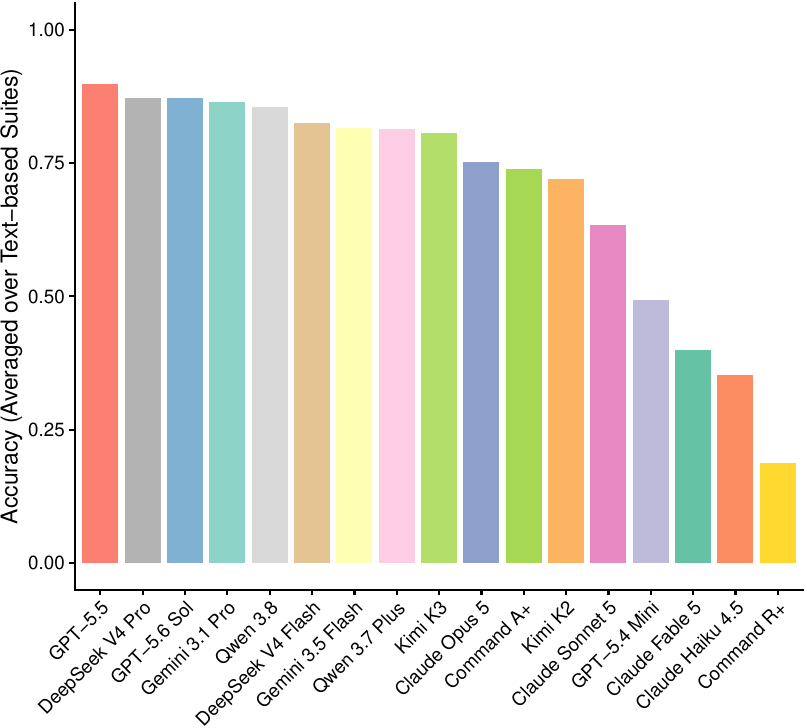}}

}

\caption{\label{fig-overall-accuracy-by-model-text}Performance by model,
on only text-based suites.}

\end{figure}%

\newpage

\section{Trial Run (July 20, 2026-July 28, 2026)}\label{sec-appendix-c}

\subsection{Accuracy over time}\label{accuracy-over-time-1}

\begin{figure}

\centering{

\pandocbounded{\includegraphics[keepaspectratio]{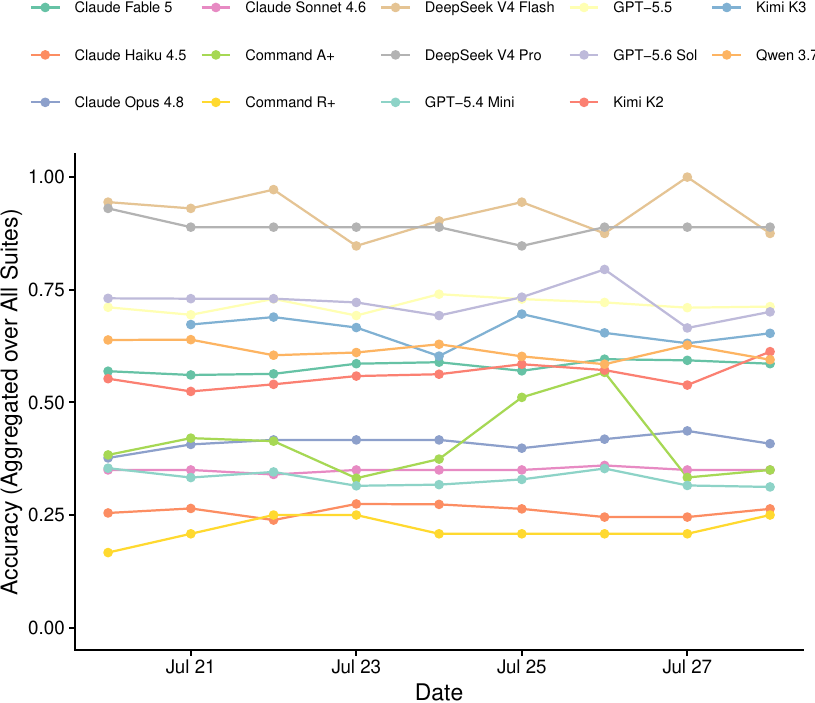}}

}

\caption{\label{fig-model-accuracy-over-time-test}Time series of model
performances from July 20, 2026 to July 28, 2026. Model performance is
the average accuracy scored by the model across all suites. We note that
DeepSeek V4 Flash, DeepSeek V4 Pro, and Command R+ displayed in this
Figure were only evaluated on a subset of suites.}

\end{figure}

Average daily accuracy for each model from July 20, 2026 to July 28,
2026 is displayed in Figure~\ref{fig-model-accuracy-over-time-test}. All
models are shown. For all models, performance appears to fluctuate
slightly from day to day. Some models like Claude Sonnet 4.6, Claude
Opus 4.8 and Kimi K2 have relatively stable performances over time, with
only mild deviations from their series average. Others like Command A+,
GPT-5.6 Sol and Qwen 3.7 Plus have much more variable performance over
time. These fluctuations were substantial enough to affect the rankings
of the models. For instance, on July 27, Command A+ performed worse than
Claude Haiku 4.5 and GPT-5.4 Mini, when it had performed better than
both of them the day before.

On July 26, GPT-5.6 Sol achieved an all-time high performance of almost
75 percent, after being outperformed by GPT-5.5 and Claude Fable 5, just
days before and being outperformed again the day after. This
demonstrates the variability in model performance on the same tasks over
time, supporting the importance of capturing and reporting daily
evaluations. On July 29, OpenAI announced on Twitter/X that GPT-5.6 Sol
was altered post-deployment to refine performance.

\begin{figure}

\centering{

\includegraphics[width=5.75in,height=\textheight,keepaspectratio]{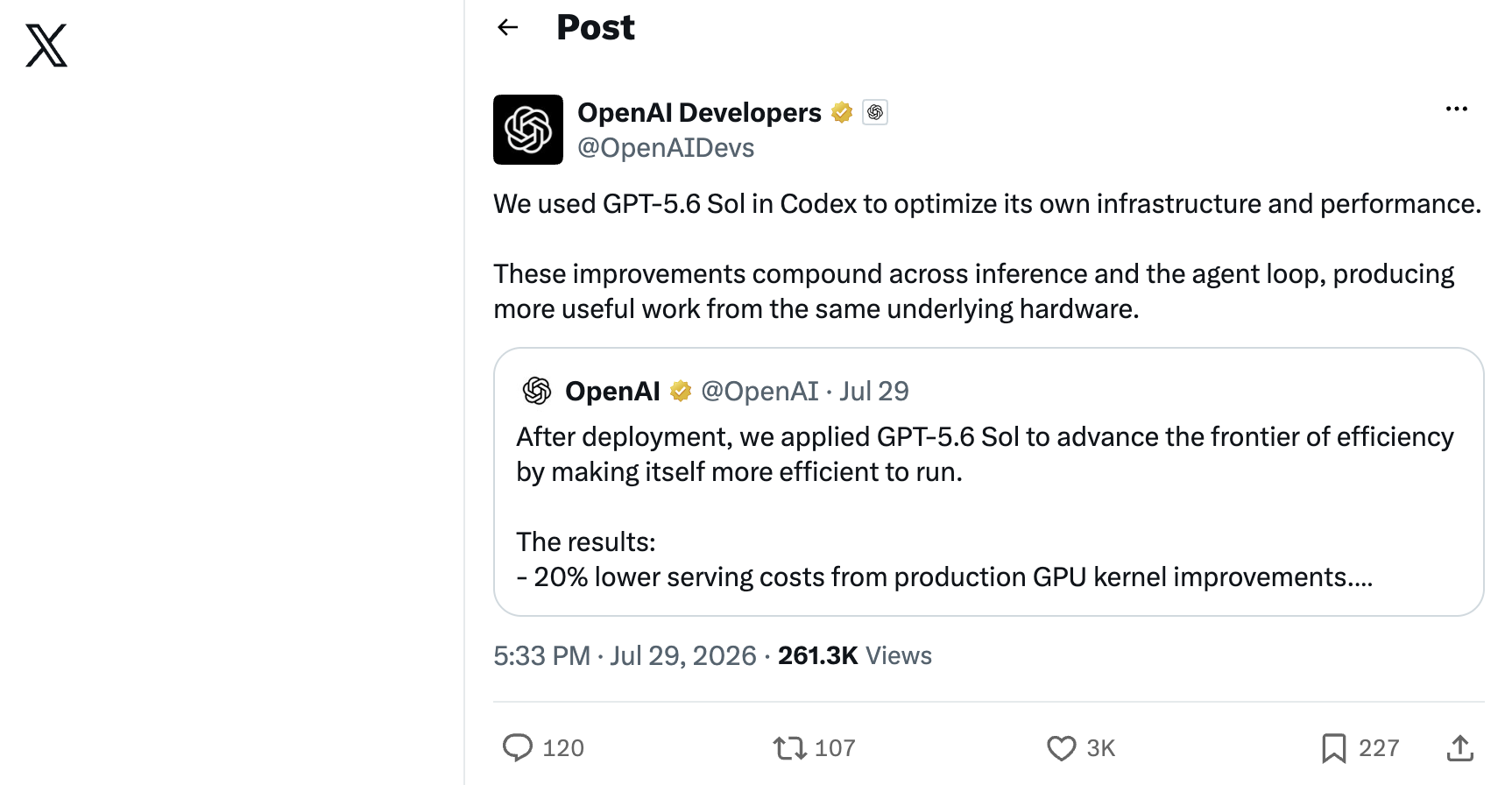}

}

\caption{\label{fig-openai-twitter-post}Public post on Twitter/X by
OpenAI developers disclosing new post-deployment enhancements made to
GPT-5.6 Sol.}

\end{figure}%

\newpage

\section*{References}\label{references}
\addcontentsline{toc}{section}{References}

\protect\phantomsection\label{refs}

\end{document}